\documentclass[letterpaper,11pt,twocolumn]{article}

\newcommand{\reportnumber}{TR-RoboPI-2603}
\newcommand{\reportdate}{November 07, 2026} % or \today
\newcommand{\shorttitle}{HMI for Subsea Telerobotics}
\newcommand{\fulltitle}{Human-Machine Interfaces for Subsea Telerobotics:\\ From Soda-straw to Natural Language Interactions}

\usepackage[letterpaper, top=1in, bottom=1in, left=0.75in, right=0.75in,
            headheight=14pt, headsep=14pt]{geometry}
\usepackage[T1]{fontenc}
\usepackage{lmodern}
\usepackage{microtype}

\usepackage{xspace}
\usepackage[table]{xcolor}
\definecolor{UFblue}{HTML}{0021A5}
\definecolor{UForange}{HTML}{FA4616}
\definecolor{lightgray}{gray}{0.95}

\usepackage{cite}
\usepackage{amsmath,amssymb,amsfonts}
\usepackage{amsthm}
\theoremstyle{plain}

\theoremstyle{remark}

\usepackage{graphicx}
\usepackage{gensymb,pifont}
\usepackage[size=small, labelfont={bf,sf}, labelsep=period]{caption}
\usepackage{subcaption}
\usepackage{booktabs}
\usepackage{tabularx}
\usepackage{multicol}
\usepackage{multirow}
\usepackage{soul}
\usepackage{siunitx}
\usepackage{physunits}
\usepackage{algorithm}
\usepackage{algorithmic}
\usepackage{array}
\usepackage[normalem]{ulem}
\usepackage{makecell}
\usepackage{wrapfig}
\usepackage{lipsum}
\usepackage{ragged2e}

\usepackage[shortlabels]{enumitem}

\usepackage{titlesec}
\titleformat{\section}
  {\normalfont\large\sffamily\bfseries\color{UFblue}}{\thesection}{0.6em}{}
\titleformat{\subsection}
  {\normalfont\footnotesize\sffamily\bfseries\color{UFblue}}{\thesubsection}{0.5em}{}
\titleformat{\subsubsection}
  {\normalfont\footnotesize\sffamily\itshape}{\thesubsubsection}{0.5em}{}
\titlespacing*{\section}{0pt}{2.0ex plus 0.5ex minus 0.2ex}{1.0ex plus 0.2ex}
\titlespacing*{\subsection}{0pt}{1.5ex plus 0.5ex minus 0.2ex}{0.6ex plus 0.2ex}
\titlespacing*{\subsubsection}{0pt}{1.2ex plus 0.3ex minus 0.2ex}{0.4ex}

\usepackage{fancyhdr}
\usepackage{xurl}
\makeatletter
\let\NAT@parse\undefined
\makeatother
\usepackage[colorlinks=true, linkcolor=UFblue, citecolor=UFblue,
            urlcolor=UForange]{hyperref}
\makeatletter
\DeclareRobustCommand\onedot{\futurelet\@let@token\@onedot}
\def\@onedot{\ifx\@let@token.\else.\null\fi\xspace}
\def\eg{\emph{e.g}\onedot} 
\def\ie{\emph{i.e}\onedot} 
 
\def\etc{\emph{etc}\onedot} 
 
\def\etal{\emph{et al}\onedot}
\makeatother
\long\def\invis#1{}

\newcommand\rebuttal[1]{{#1}}
\newcommand\finalrev[1]{{#1}}

\long\def\invis#1{}
\newcommand{\tick}{\ding{51}} % Tick 
\newcommand{\cross}{\ding{53}} % Cross 

\usepackage{times}
\hypersetup{
  pdftitle={\fulltitle},
  pdfauthor={Adnan Abdullah, David Blow, Ruo Chen, Thanakon Uthai, Eric Jing Du, Md Jahidul Islam},
  pdfsubject={\reportnumber}
}

\begin{document}

% TITLE PAGE (single column, no page number)

\begin{titlepage}
\centering
\sffamily

% --- Top banner ---
{\color{UFblue}\rule{\textwidth}{2pt}}\par\vspace{4pt}
\begin{minipage}{0.5\textwidth}
  \raggedright
  {\large\bfseries\color{UFblue} TECHNICAL REPORT}
\end{minipage}%
\begin{minipage}{0.5\textwidth}
  \raggedleft
  {\small\bfseries Report No.\ \reportnumber}\\[2pt]
  {\small \reportdate}
\end{minipage}\par
\vspace{4pt}{\color{UFblue}\rule{\textwidth}{0.6pt}}\par

\vspace{0.6in}

% --- Title ---
{\Large\bfseries\color{UFblue} \fulltitle \par}

\vspace{0.3in}
{\color{UForange}\rule{0.25\textwidth}{1.5pt}}\par
\vspace{0.3in}

% --- Authors ---
{\large
Adnan Abdullah$^{1}$, David Blow$^{1}$, Ruo Chen$^{1}$, Thanakon Uthai$^{2}$,\\ Eric Jing Du$^{2}$, and Md Jahidul Islam$^{1}$\par}
\vspace{0.15in}
{\normalsize\color{gray}
$^{1}$RoboPI Laboratory, University of Florida\\
$^{2}$ICIC Laboratory, University of Florida\par}

\vspace{0.35in}

% --- Report version information ---
{\normalsize\bfseries\color{gray}
First Published: Dec 2024\par}

\vspace{3pt}
{\normalsize\color{gray}
Revised: Nov 2026\par}

\vspace{0.2in}

% --- Revision note ---
\begin{minipage}{0.88\textwidth}
\rmfamily\small
% \textbf{Revision note:}
% Originally released as an arXiv preprint in 2025. Updated as technical report in November 2026.
\end{minipage}

\vspace{0.35in}

% --- Abstract box ---

% \vspace{0.6in}

% --- Abstract box ---
\begin{minipage}{0.88\textwidth}
\colorbox{lightgray}{%
\begin{minipage}{\dimexpr\textwidth-2\fboxsep\relax}
\vspace{6pt}
\rmfamily\small
{\sffamily\bfseries\color{UFblue} Abstract}\par\vspace{4pt}
\justifying
This review explores the evolution of human-machine interfaces (HMIs) in subsea telerobotics, charting the progression from traditional first-person ``soda-straw" consoles-- characterized by narrow field-of-view camera feeds-- to contemporary interfaces leveraging gesture recognition, virtual reality, and natural language processing. The review makes three key contributions. First, we introduce a structured framework that analyzes subsea HMIs through three interrelated aspects: operator experience, robotic autonomy, and bidirectional human-machine communication. Second, using these aspects, we compare visual, haptic, extended-reality, gesture, and natural-language interfaces; examine the progression from direct teleoperation to shared autonomy; and assess the distinct roles of simulators and digital twins in training, prototyping, and mission support. Third, we synthesize the reported evidence to distinguish established industrial tools, selectively deployed technologies, and research-stage capabilities, while identifying limitations in interoperability, communication reliability, immersive feedback, and field validation. From this synthesis, we outline standardization needs and research directions for hydrodynamic feedback, latency-aware shared autonomy, grounded language interaction, and real-to-simulation transfer. The report concludes by identifying open research questions and proposing a roadmap toward more intelligent and intuitive subsea HMI systems.

\vspace{6pt}
{\sffamily\bfseries\color{UFblue} Keywords:}
Human-machine interface; subsea telerobotics.
\vspace{6pt}
\end{minipage}}
\end{minipage}

\vfill

% --- Footer of title page ---
{\color{UFblue}\rule{\textwidth}{0.6pt}}\par\vspace{4pt}
{\small\color{gray} University of Florida \,$\cdot$\, Gainesville, FL, USA}
\end{titlepage}

% MAIN BODY (two columns, numbering restarts at page 1)

\vspace{-2mm}
\section{Introduction}\label{intro}
Subsea telerobotic technologies have advanced significantly over the past few decades, driven by the need for remote inspection, maintenance, and research expeditions in underwater environments~\cite{kunz2008deep,ryden2013advanced}. Remotely operated vehicles (ROVs) and autonomous underwater vehicles (AUVs) are typically deployed to perform remote tasks in subsea environments that are beyond the reach of human scuba divers~\cite{petillot2019underwater}. The robots are generally teleoperated from a surface vessel or a base station~\cite{lin1997virtual} for applications such as underwater infrastructure inspection~\cite{jacobi2015autonomous,patel2024multi,islam2024eob}, seabed mapping~\cite{joe2021sensor,howard2006experiments}, remote surveillance~\cite{terracciano2020marine,ferri2020cooperative}, environmental monitoring~\cite{teague2018potential}, scientific expeditions, and more. \rebuttal{Despite growing interest in autonomy, teleoperation continues to dominate subsea missions, comprising over 75\% of deployments~\cite{GrandView}. This trend highlights the immense potential of advancing telerobotic interfaces for economic growth and scientific discovery.}
% The National Oceanic and Atmospheric Administration (NOAA) estimates that approximately $95\%$ of the world's oceans and $99\%$ of the ocean floor are unexplored by human~\cite{xia2023sensory}. 
% Enhancing telerobotics capabilities in subsea engineering tasks thus presents a significant opportunity for economic growth and scientific discovery.

Early teleoperated systems were limited in terms of the feedback they provided, often relying on basic visual data from low-resolution cameras. The operator had to rely heavily on experience and intuition to navigate and perform tasks, which increased the cognitive load and risk of errors. While effective for simpler tasks, studies indicate that the limited field-of-view (FOV), appearing as looking through a ``\textbf{\textit{soda straw}}''~\cite{abdullah2024ego2exo}, results in reduced situational awareness in more complex and dynamic environments, leading to increased cognitive load and reduced performance. The 2D visual feedback provides only a partial understanding of the remote workspace, requiring operators to frequently switch between different viewpoints if available~\cite{rahnamaei2014automatic}, and manually integrate information from various perspectives. 
\rebuttal{
This challenge is especially pronounced in confined or cluttered spaces, a common scenario in underwater pipelines, caves, and shipwreck exploration. In larger workspaces such as coral reefs, offshore rigs, and mining fields, traditional monocular camera-based interfaces fail to provide a comprehensive spatial view, leading to significant cognitive overload for teleoperators~\cite{xie2022underwater,elor2021catching}.
}

\begin{figure*}[t]
    \centering
    \includegraphics[width=0.97\linewidth]{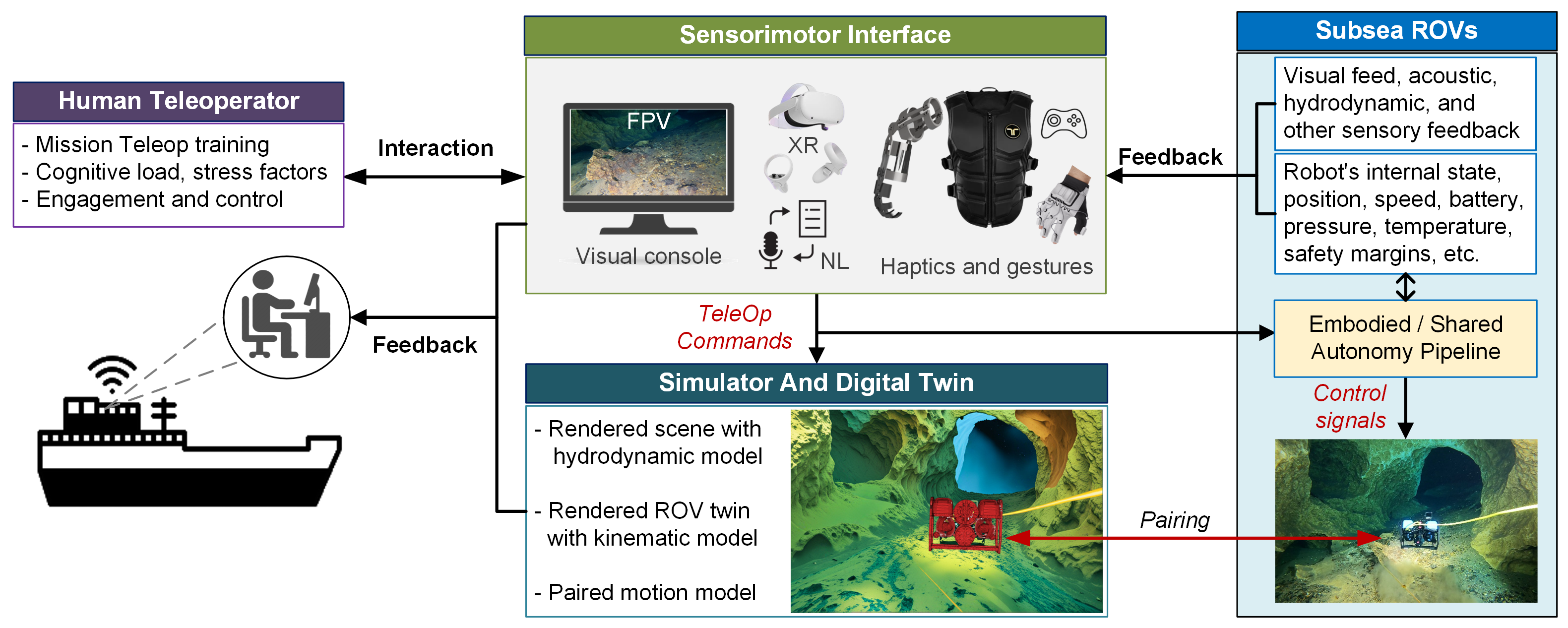}%
    \vspace{-2mm}
    \caption{An outline of the human-robot sensorimotor interaction flow for subsea telerobotics is shown. The interactions occur via control and feedback devices such as visual consoles, haptic gloves or suits, HMDs, and natural language interfaces. A virtual simulation engine is often integrated with the interface to project the remote environment and ROV operation as a \textit{digital twin}. Simulators with integrated twin models help teleoperation training by mission rehearsal/replay, which is essential for analyzing teleoperators' cognitive load as well as for prototyping human-machine embodied/shared autonomy features.  (FPV: first-person view; NL: natural language; XR: augmented/virtual/mixed reality)
    }
    \label{fig:overview}
    \vspace{-4mm}
\end{figure*}

\finalrev{These challenges motivated 3D augmented/ virtual/ mixed reality (XR) visualization, haptic feedback, natural language (NL) control, and shared-autonomy interfaces. Interface design directly affects operators' situational awareness, workload, stress, and task performance~\cite{van2019human,riley2016situation,lathan2002effects}. As summarized in Fig.~\ref{fig:overview}, the HMI couples operator commands with robot and environmental feedback; simulation and digital models can further augment visualization and training~\cite{manhaes2016uuv}, while shared autonomy delegates selected low-level functions without removing human oversight~\cite{lee2016development,xia2024human,phung2024shared}.}

\vspace{0.5mm}
\noindent
\rebuttal{
\textbf{Knowledge gaps and open questions}. 
The interdisciplinary nature of modern HMIs with haptics, XR, and NL features has led to a broad but fragmented body of literature. The existing systems and interfaces are built for a diverse set of platforms and applications without any consistent guidelines. \finalrev{System reports focus on individual technologies, with limited assessment of deployment maturity. Moreover, there is no comprehensive technical review that compares the interfaces, compiles the findings, and identifies integration opportunities.} To this end, we pose the following research questions, highlighting key gaps in HMI design, communication modalities, interaction strategies, and autonomy frameworks.
\begin{enumerate}[left=0pt,nolistsep]
    \item What design principles and standard guidelines are necessary to ensure modularity and cross-platform compatibility of HMIs in subsea telerobotics? %\JI{which section of the paper answers this?} \Adnan{These are open research questions to explore. We talk about guidelines in Sec.-II but do not address the problem.}
    \item What command and feedback modalities (\eg, language, gesture, haptics, vision) should be prioritized or fused for different tasks according to complexity, environmental constraints, and mission objectives? %\JI{need to revise}
    \item To what extent can digital replicas (simulators and twins) mimic a real-world scenario, thereby improving operator skills, reducing mission logistics, and supporting algorithm development for subsea operations?
    \item \finalrev{How can shared-autonomy mechanisms be integrated into modern interfaces and adapted to the operator's skill level and cognitive load?}
    %\JI{does this paper talk about `machine intelligence' ? various skill levels etc? which section answers this?} \Adnan{Open question. But we show examples in Sec.-V, Table-IV}
\end{enumerate}
An in-depth understanding of these aspects is essential for driving innovation and advancing the HMI frontiers of subsea telerobotics. This report aims to contribute toward that objective by providing a structured technical review of the key HMI dimensions relevant to marine robotic systems.
}

% \Adnan{Flow: Gap - RQ - motivation - topics of this paper - evaluation guidelines [hands-on exp, from other eval, from search] - contri - how readers would be benefitted}

% this review attempts not only to compile the vastly scattered knowledge of subsea human-machine interfaces (HMIs) but also to provide a structured, evaluative lens through which state-of-the-art systems can be compared. The paper is written from the perspective of researchers actively engaged in subsea robotics and aims to bridge the gap between academic progress and practical mission needs.

\vspace{1mm}
\noindent
\rebuttal{
\textbf{Specific contributions}. 
In this report, we present a structured analysis of subsea HMI systems based on their functional roles, interaction modalities, and levels of autonomy as observed in the literature. To facilitate comparative evaluation, we introduce the concept of ``\textit{technology clusters}'' that delineate how human-machine interaction happens across different system architectures. First, we categorize and evaluate the state-of-the-art (SOTA) subsea telerobotics interfaces and simulators based on a set of guidelines such as their choice of sensory modalities, platform compatibility, user comfort, and optimal use cases (Section~\ref{sec:interfaces}-\ref{sec:simulator}). Then, we present a detailed review of human-robot teaming in subsea telerobotics, highlighting various aspects of shared autonomy, bidirectional interaction, and intelligence (Section~\ref{sec:autonomy}). Finally, we analyze the SOTA approaches and trends to address the challenges in subsea HMIs, highlighting open problems and use cases for future research (Section~\ref{sec:discussion}).
}

\rebuttal{
This review is authored from the perspective of researchers actively involved in subsea robotics to bridge the gap between academic advancements and the practical demands of real-world mission scenarios. It aims to provide foundational knowledge to students, researchers, scientists, and engineers engaged in HMI development. The classification and open research questions will offer a roadmap to young researchers for systematic investigation into interface design, autonomy integration, and human-robot collaboration. For industry leaders, the comparative insights into current systems and emerging trends will inform the design of futuristic HMI solutions. 
}

%\JI{This paragraph is too shallow and full of cheap sentences. Need to revise}

% \reviewer{(\#3)
% \begin{itemize}
% \item Are there any emerging efforts or industry initiatives towards standardization that could be mentioned? Is it required? Is it necessary? Some emphasis could be given here. Perhaps some insight on the paper?
% \item What specific aspects of subsea HMI design would benefit most from standardization (e.g., data formats, control protocols, user interface elements)? 
% \item Can you suggest potential directions for developing future guidelines or standards in this area?
% \end{itemize}
% }

\begin{figure*}[t]
    \centering
    \includegraphics[width=0.95\linewidth]{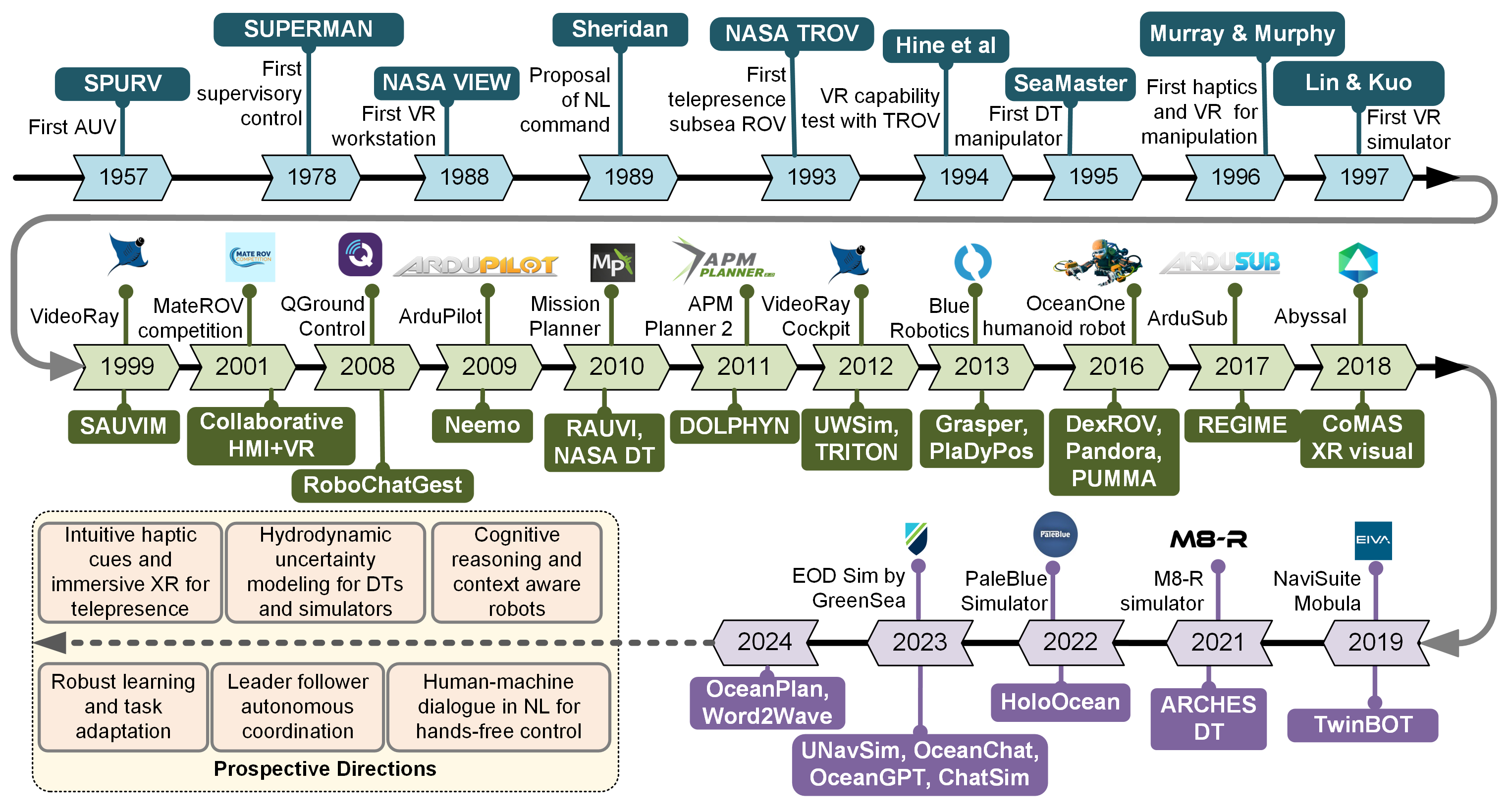}%
    \vspace{-2mm}
    \caption{A chronological evolution of subsea HMI technologies is shown. The top row highlights early achievements and milestones. As technology advanced, this century saw industry solutions for interfaces and simulators, incorporating XR and haptic features (second row). In the last five years, NL-based interactions have gained popularity, driven by the success of LLMs.%
    }
    \label{fig:tech_evolution}
    \vspace{-2mm}
\end{figure*}

\section{Review Scope And Criteria}\label{sec:scope}
\rebuttal{
This review spans two primary HMI categories: commercially available interfaces with publicly accessible documentation and peer-reviewed interfaces within the academic literature. Although current efforts toward HMI standardization emphasize the adoption of open-source, widely used, and accessible platforms-- such as BlueRobotics hardware~\cite{bluerov}, Robot Operating System (ROS2) middleware~\cite{quigley2009ros}, and ArduPilot-based autopilot architectures~\cite{ardusub}-- these initiatives largely reflect localized, predominantly U.S.-based community practices. As such, they do not constitute benchmarks that are consistently adopted across the academic and industrial landscapes.
}

\finalrev{We sort literature through targeted searches of IEEE Xplore, Scopus, Web of Science, and Google Scholar, supplemented by cross-referencing, from foundational studies through $2026$. Search terms include ``underwater,'' ``subsea,'' ``marine,'' ``ROV,'' or ``AUV'' with ``human-machine interface,'' ``teleoperation,'' ``XR,'' ``haptic,'' ``natural language,'' ``shared autonomy,'' ``digital twin,'' or ``simulator.'' Title and abstract screening is followed by full-text assessment. We include works describing interface design, human interaction, remote control, or relevant evaluation, and exclude studies limited to perception or autonomy without an HMI connection. Peer-reviewed sources support technical claims; official documentation and demonstrations are used for product features and are marked accordingly. Representative works are highlighted if they introduce a distinct interaction modality, system capability, or evaluation result; nevertheless, lack of commercial disclosure means that the coverage is not exhaustive and may favor well-documented systems.}

\rebuttal{
Given the lack of established benchmarks in this domain, we introduce evaluation criteria grounded in three core aspects: \textbf{(i)} human operators' cognitive experience with feedback and control modalities, \textbf{(ii)} the robot's degree of autonomy and its relevance for various tasks, and \textbf{(iii)} the quality of bidirectional human-robot communication. \finalrev{The three aspects reflect the principal components of closed-loop telerobotic interaction: the operator perceives remote feedback and issues commands, the robot interprets and executes those commands, and the communication and authority-allocation mechanisms couple the two~\cite{zhou2020intuitive}. Accordingly, Sec.~\ref{sec:interfaces} examines operator-side command and feedback modalities, Sec.~\ref{sec:simulator} discusses simulation support, Sec.~\ref{sec:autonomy} analyzes robot-side task execution, and Sec.~\ref{sec:discussion} addresses the cross-cutting communication, modeling, and validation constraints.} Detailed criteria for each evaluation and analysis are presented in the respective sections; itemized features are marked using the following symbols for normalized interpretation:
\begin{itemize}[nosep,leftmargin=8pt,label={}]
    \item $\dagger$: ~Supported by peer review or official documentation
    \item $\ddagger$: ~Derived from demo/promotional video
    \item $\circledast$: Reported in user studies
    \item $\blacklozenge$: ~Inferred by the authors based on others' findings; should be treated as illustrative insights rather than definitive claims.
\end{itemize}
}

This approach, while not exhaustive, reflects the broader need for formal standardization efforts in HMI technologies. \finalrev{In Tables~\ref{tab:interface} and~\ref{tab:haptic_interfaces}, factual features are reported only when supported by peer-reviewed studies, official documentation, or demonstrations; inferred use cases and ergonomic assessments are explicitly marked and should not be interpreted as measured performance. Beyond consistent grouping, standardization can improve legacy-system interoperability, particularly at the data and interface layers. Middleware adapters or protocol gateways can translate platform-specific telemetry, commands, sensor streams, and mission logs into common formats and APIs without replacing existing hardware or low-level controllers. Closed protocols, limited onboard computation, and safety-critical interfaces may nevertheless require vendor-specific modification~\cite{kabanov2022marine,johansson2021maritime}.}

\section{Subsea Telerobotic User Interfaces} \label{sec:interfaces}
% \vspace{-1mm}

\finalrev{Subsea HMIs have progressed by addressing limitations exposed by their earlier generations. Camera-based control enabled reliable remote operation but imposed a narrow field of view and high workload~\cite{lin1997virtual}; haptic and XR interfaces added physical and spatial feedback~\cite{dennerlein2000vibrotactile,hine1994application,wang2024robotic}; and shared autonomy reduced continuous low-level control under workload and communication constraints. Recent NL interfaces shift interaction toward high-level mission management~\cite{yang2024oceanplan}, but they leave unresolved requirements for grounded interpretation, error handling, and safe authority allocation.}

% Telerobotic interfaces have evolved significantly over the past decades, with novel integrated features playing a critical role in enhancing operator control and mission success rate. Traditionally, vision-based interfaces dominated, where operators relied solely on 2D video feeds to navigate and control a robot~\cite{lin1997virtual}. Haptic feedback was integrated later~\cite{dennerlein2000vibrotactile}, improving task precision by allowing operators to feel forces such as drag and resistance. The introduction of XR interfaces further advanced teleoperation by immersing operators in a 3D virtual environment, providing greater situational awareness~\cite{hine1994application,wang2024robotic}. More recently, researchers have integrated natural language understanding capability into interfaces~\cite{yang2024oceanplan}, allowing operators to plan missions and control ROVs using NL commands, making interaction more intuitive. 
% Fig.~\ref{fig:tech_evolution} summarizes the early milestones in marine telerobotics, their progress over the past six decades, SOTA interface products, and potential future innovations.

\subsection{Status Quo of Robot Teleoperation Interfaces}
%\reviewer{(\#5) classification of features for the existing subsea Teleop interfaces - and for the evolutionary map - is not clearly declared.}

Subsea HMIs have been developed to address the unique challenges of underwater operations, where visibility, communication bandwidth, localization techniques, and environmental conditions significantly differ from terrestrial systems. 
\finalrev{Fig.~\ref{fig:tech_evolution} groups subsea HMI development into three eras: foundational space and ocean exploration, the emergence of commercial marine-robotic interfaces, and the recent adoption of immersive and language-driven interaction. These eras illustrate the joint evolution of technical capability, commercial availability, and operator requirements.}

\finalrev{A teleoperation interface is commonly implemented as a ground control station (GCS)~\cite{ardupilotGCS}. Examples include QGroundControl, APM Planner2, EOD Workspace, and NaviSuite Mobula, together with mobile tools such as DroidPlanner, SidePilot, and MAVPilot~\cite{qgroundcontrol,apmplanner2,eod,navisuite,droidplanner,sidepilot,mavproxy}. Some mobile interfaces were developed for GPS-dependent aerial vehicles and are therefore not directly transferable underwater. Table~\ref{tab:interface} compares the interfaces using three groups of criteria: \textbf{(i)} accessibility and integration, including source availability and operating-system support; \textbf{(ii)} operational support, including recording, waypoint navigation, diagnosis, replay, and multi-vehicle operation; and \textbf{(iii)} situational awareness through mapping and 3D visualization. The comparison is necessarily incomplete where commercial capabilities are not publicly disclosed.}

\finalrev{Since feature availability alone does not indicate operational value, we consider two interpretations: platform-level trade-offs and technology maturity. The surveyed interfaces reveal a practical trade-off between general-purpose, open platforms and vertically integrated proprietary systems. Open-source interfaces such as QGroundControl emphasize accessibility, making them suitable for research and heterogeneous robots. In contrast, proprietary interfaces integrate vehicle-specific visualization, diagnostics, mission replay, or autonomous functions, limiting portability across platforms. Thus, interface selection depends not only on the number of available features, but also on whether the application prioritizes customization and interoperability or mission-specific operation.}

\vspace{1mm}
\noindent
\finalrev{\textbf{Technology maturity.} The survey spans both commercial systems and research prototypes, whose capabilities are not equally mature. Therefore, we group them into three categories. \textit{Established industrial functions} include camera/joystick control, telemetry, video recording, waypoint navigation, map overlays, and fault monitoring, which are routinely available in commercial ground-control systems~\cite{qgroundcontrol,videoray-cockpit,navisuite}. \textit{Selectively deployed technologies} include XR visualization, mission replay, and supervisory control; their broader adoption remains constrained by platform-specific integration, cost, training requirements, and poor cross-vendor interoperability~\cite{abyssal_offshore,gancet2015dexrov}. \textit{Research-stage capabilities} include full-body or hydrodynamic haptics, NL-based mission execution, and adaptive or learned autonomy~\cite{xia2023rov,yang2024oceanplan,lawrance2019shared}. Their transition to practice requires reliable sensing and communication, broader field validation, safe authority transitions, and appropriate certification.}
% Among open-source interfaces, QGroundControl is the most popular across the ArduPilot community~\cite{ardupilot}. It is built on ArduSub~\cite{ardusub}, a comprehensive open-source framework specifically designed for underwater vehicles. Among proprietary interfaces, leading ROV manufacturer VideoRay offers VideoRay Cockpit~\cite{videoray-cockpit}, exclusively designed to operate their ROVs. NaviSuite Mobula is a third-party interface built for VideoRay ROVs and offers some autonomous capabilities including scanning a wall, orbiting a target, \etc. Abyssal Offshore~\cite{abyssal_offshore} integrates real-time XR visualization, $270\degree$ FOV from multiple cameras, and risk analyses for commercial ROV operation. 
% Open-source platforms such as QGroundControl are more suitable for researchers since they allow customization to suit specific hardware configurations and operational needs. 
% OpenROV cockpit~\cite{openrov-cockpit} offers browser-based connectivity without requiring any software installation. Industry-grade interfaces such as FMC Schilling~\cite{gri} and Teledyne SeaBotix~\cite{teledyne_seabotix} come as rugged hardware packages and do not require external computers or controllers for teleoperation.

\begin{table*}[t]
    \centering
    \caption{Comparison of existing subsea teleop interfaces based on their feature and platform compatibility. ($\infty$: unlimited number; W: Windows; M: MacOS; L: Linux); other notations are explained in Sec.~\ref{sec:scope}. %\reviewer{(\#6) It is not clear if the features of table are not available in the standard version, or are customizable. It is not simple to state the completeness of table I in terms of size and in term of native subsea teleop interfaces (this could be a relevant discussion point)}
    }
    \vspace{-1mm}
    \resizebox{\textwidth}{!}{
    \begin{tabular}{lccccccccc}
     \Xhline{2\arrayrulewidth}
    \rowcolor[rgb]{0.9,0.9,0.9}
       & Open & Video & Waypoint & Scene & 3D Visual  & Fault & Mission & Max. \# & Supported \\
       \rowcolor[rgb]{0.9,0.9,0.9}& Source$^\dagger$ & Recorder$^\dagger$ & Navigation$^\dagger$ & Mapper$^\ddagger$ & Supported$^\ddagger$  & Diagnosis$^\ddagger$ & Replay$^\dagger$ & Vehicles$^\dagger$ & {OS: W,M,L}$^\dagger$ \\
      \Xhline{2\arrayrulewidth}
      QGroundControl~\cite{qgroundcontrol} & \tick & \tick & \tick & \cross & \cross  & \tick & \cross & $7$ & \tick \tick \tick \\
      Mission Planner~\cite{missionplanner} & \tick & \cross & \tick & \cross & \cross & \tick & \cross & $\infty$ & \tick \tick \cross \\
      APM Planner2~\cite{apmplanner2} & \tick & \tick & \tick & \tick & \cross & \tick & \cross & $\infty$ & \tick \tick \tick \\
      % UgCS~\cite{ugcs} & \cross & \tick & \tick & \tick & \tick & \tick & \tick & $\infty$ & \tick \tick \tick \\
      EOD Workspace~\cite{eod} & \cross & \tick & \tick & \tick & \cross & \tick & \cross & $\infty$ & \tick \cross \cross \\
      VideoRay Cockpit~\cite{videoray-cockpit} & \cross & \tick & \cross & \cross & \tick & \tick & \tick & $1$ & \tick \cross \cross \\
      Teledyne SeaBotix~\cite{teledyne_seabotix} & \cross & \tick & \tick & \cross & \cross & \tick & \cross & $1$ & \tick \cross \cross \\
      SRS Fusion~\cite{srs_fusion} & \cross & \tick & \tick & \tick & \tick & \cross & \tick & $1$ & \tick \cross \cross \\
      NaviSuite Mobula~\cite{navisuite} & \cross & \tick & \tick & \tick & \tick & \cross & \tick & $1$ & \tick \cross \cross \\
      Abyssal Offshore~\cite{abyssal_offshore} & \cross & \tick & \cross & \cross & \tick & \tick & \tick & $1$ & \tick \cross \cross \\
      BRIDGE~\cite{bridge} & \cross & \tick & \tick & \tick & \cross & \cross & \tick & $1$ & \tick \cross \cross \\
      SubNav~\cite{subnav} & \cross & \tick & \tick & \cross & \cross & \tick & \tick & $1$ & \tick \tick \tick \\
      DroidPlanner~\cite{droidplanner} & \tick & \cross & \tick & \cross & \cross & \cross & \cross & $1$ & {Android} \\
      \bottomrule
    \end{tabular}
    }
    \label{tab:interface}
    \vspace{-3mm}
\end{table*}

\subsection{Vision-based Consoles}
Underwater vision is a well-explored area in telerobotics literature and has long been the key navigational information for teleoperators~\cite{islam2024computer}.
% , as 90\% of human perception rely on vision~\cite{lichiardopol2007survey}. 
The environmental challenges as well as the inherent limitations of 2D visual systems have been studied for decades, and many innovative solutions have been proposed. We summarize the following challenges and corresponding countermeasures employed in vision-based interfaces. First, traditional consoles with a fixed monocular POV lack depth perception and fail to provide peripheral vision, resulting in reduced situational awareness. Second, low-light turbid water conditions reduce visibility and hinder the capture of clear images. Third, the ROV's onboard lights get reflected and back-scattered by suspended particles directly at the front camera, creating glare and large blind spots for the operator~\cite{yu2023weakly}. 

To compensate for poor visibility and to provide the operator with a comprehensive understanding of the scene, modern marine telerobotic systems use high-definition wide-angle cameras~\cite{singh2007towards}, structured lighting systems~\cite{jaffe2010enhanced}, image enhancement~\cite{islam2020fast}, and scene parsing technologies~\cite{abdullah2023caveseg}. However, interpreting wide-angle fisheye camera images on 2D displays is difficult because of the distortion effects they introduce. Instead, contemporary works generate third-person-views using augmented reality~\cite{abdullah2024ego2exo}, follower ROV~\cite{nagatani2011redesign}, external camera~\cite{sato2013spatio}, \etc.
% Although not popular yet, multispectral and hyperspectral imaging has the potential to provide more robust vision systems with enhanced marine life detection and mapping mechanisms~\cite{liu2020underwater}. 
While vision will remain the fundamental modality for teleoperation, other sensory augmentations are essential for the immersive realization of the remote environment~\cite{xia2023sensory}.

% Besides the ROV's front camera feed, the visual consoles display a wide range of internal and external sensor readings \ie, telemetry data\invis{in terms of logs, graphs, \etc which are crucial for safe navigation}. External sensors capture information such as pressure, depth, temperature, water current, acoustic scans, \etc. Researchers emphasize scanning sonar systems (\eg side scan sonar, synthetic aperture sonar) for safer navigation, as they detect and display obstacles beyond the limited FOV of cameras~\cite{choi2021compressive,rahman2019svin2}. Meanwhile, internal data encompasses the status of onboard electrical and mechanical systems including energy consumption, housing temperature, data recording, and streaming status, as well as the ROV’s pose and velocity. Among these, the attitude and heading reference system (AHRS) estimates and displays the robot's pose on the console which is crucial for maintaining heading and orientation in feature-deprived underwater scenes~\cite{krishnamurthy2013self}. 

\begin{table*}[t]
    \centering
    \caption{Haptics and XR capabilities of telerobotic interfaces are categorized and their use cases are highlighted. \rebuttal{Criteria are defined in Sec.~\ref{subsec:haptics}, notations are explained in Sec.~\ref{sec:scope}}.
    %\reviewer{(\#7) Without a clear classification of categories, It is not possible to prevent misclassification, missed data, completeness. This is an issue for table II. How comfort is measured? what does ergonomics means? how ideality in use case is set? is there possibility of integration between feedback location and feedback type (as for the case of exoskeleton and full body suits)?}
    }
    \vspace{-1mm}
    \resizebox{\textwidth}{!}{
    \begin{tabular}{lllllll}
    \Xhline{2\arrayrulewidth}
    \rowcolor[rgb]{0.9,0.9,0.9}Feedback & Feedback & Control & Ideal Use Cases$^\blacklozenge$ & User Comfort & Selected\\ 
    \rowcolor[rgb]{0.9,0.9,0.9}Location & Type$^\dagger$ & Mechanism$^\dagger$ &  & \& Ergonomics$^\circledast$$^\blacklozenge$ & References\\ 
    \Xhline{2\arrayrulewidth}
    Fingertips & Vibrotactile & Wearable gloves, & Grasping, collecting sample, & High & \cite{dennerlein2000vibrotactile,ryden2013advanced} \\ 
     &  & stylus & controlling valve/knob &  & \cite{le2016haptic} \\ \hline 
    Wrist,  & Force & Wearable gloves,  & Sensing drag and tension,  & Medium & \cite{wang2024robotic,stewart2016interactive,shazali2018development} \\ 
     forearm &  & exoskeleton & picking object  &  & \cite{girbes2020haptic,gancet2015dexrov} \\ \hline 
    Eye & Real/virtual  & HMD &  Inspecting infrastructure, & Medium &  \cite{hine1994application,elor2021catching,de2020preliminary}  \\ 
     &  video stream &  &  long-term exploration &  &  \cite{domingues2012human,xia2023visual,xia2023rov}  \\ \hline 
    Upper body & Tactile, force,  & Exoskeleton, &  Sensing and mitigating drift,  & Low & \cite{xia2023visual,xia2023rov,gancet2015dexrov}  \\ 
     & fluid flow & suit &  sensing velocity and current &  &  \\  \hline 
    Full body & Tactile, vestibular,  & Full body suit,  & Sensing velocity, pressure, current,   & Low & \cite{ha2015whole,khatib2016ocean,xu2021vr}  \\ 
     & force, fluid flow & actuated chair & using dexterity for manipulation  &  &  \\ 
    \bottomrule
    \end{tabular}
    }
    \vspace{-4mm}
    \label{tab:haptic_interfaces}
\end{table*}

\subsection{Integration of XR}
Virtually experiencing the depths of the ocean provides humans with an immersive experience, enhancing their perception and enabling more effective engagement with remote underwater settings.
% The XR interfaces allow human operator for immersive perception and interaction with the underwater environment. 
%Virtually experiencing the depths of ocean is demanding in tourism industry and leisure activities as well as in more professional tasks such as remote inspection and maintenance of subsea structures. 
In $1993$, NASA launched the first successful Telepresence Remotely Operated Vehicle (TROV)~\cite{hine1994application} under the Ross Sea ice near the McMurdo Science Station in Antarctica. This project demonstrated the feasibility of streaming video feeds from an ROV and rendering computer-generated graphics on a head-mounted display (HMD). Preliminary results also showcased ROV control mechanisms via head movements. However, the progress of XR has been slow in the subsea telerobotics domain, unlike its rapid growth in the gaming industry, surgical robotics, and aerial robotics. According to Zhou~\etal~\cite{zhou2020intuitive}, an XR model should be able to collect, process, transfer, and reconstruct the immersive scene model of the workspace in real-time and enable intuitive robot controls accordingly. The cost and logistics associated with long-term underwater missions and the unique environmental complexities pose challenges to achieving these criteria. 

\begin{figure}[t]
    \centering
    \includegraphics[width=\linewidth]{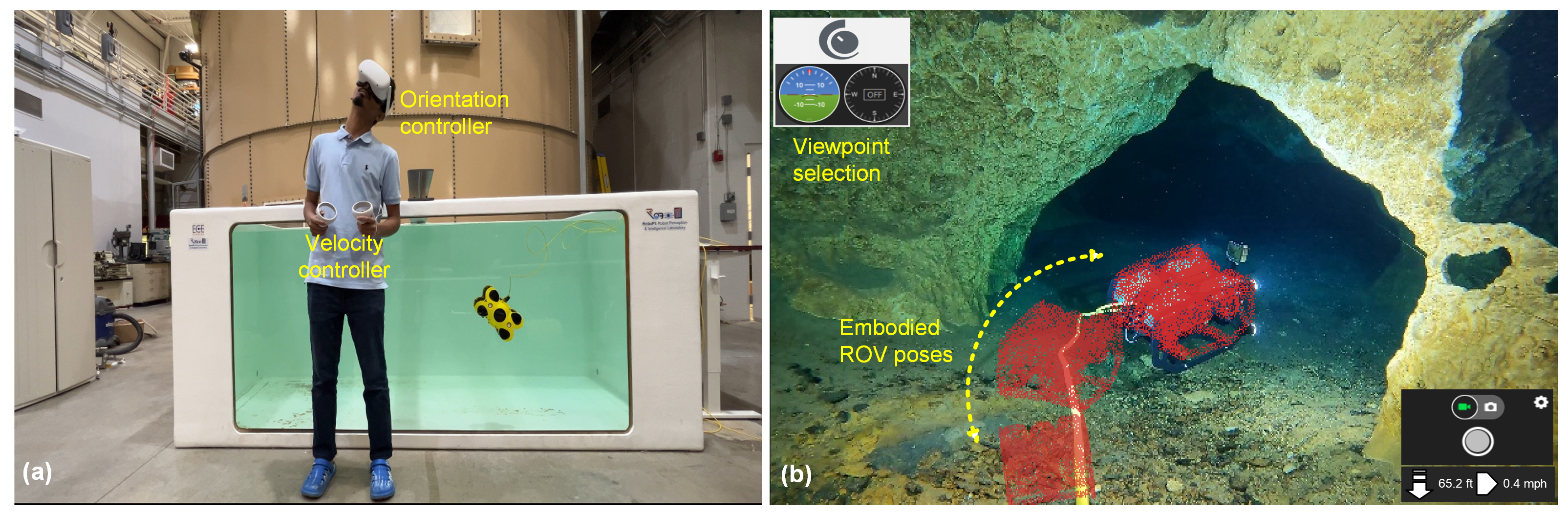}% 
    % \vspace{-2mm}
    \caption{Two telerobotics interfaces are shown; (a) An underwater ROV is being maneuvered with human head movements via HMD and haptic feedback controllers; (b) an AR interface with viewpoint augmentation capability~\cite{abdullah2024ego2exo} renders third-person EOB (eye on the back) views~\cite{islam2024eob} for peripheral information and situational awareness. 
    }
    \vspace{-4mm}
    \label{fig:haptics_xr}
\end{figure}

Researchers have come up with innovative solutions and have been successful to some extent in XR interfacing between humans and marine robots. DOLPHYN~\cite{domingues2012human}, a mixed reality-based aquatic interface for multi-sensory exploration of scuba diving sites, allows a diver to remotely operate an ROV while being aware of his own diving trajectory. Contemporary works explore XR-based view synthesis techniques to provide operators with an enhanced peripheral view. For instance, the augmented visualization of the CoMAS project~\cite{bruno2018augmented} supports ROV pilots during underwater archaeological site exploration and artifact collection. A similar approach is presented in VENUS project~\cite{chapman2008virtual}, where the archaeological content is reconstructed in mixed reality by performing bathymetric and photogrammetric surveys on the real
site. Moreover, Ego-to-Exo and EgoExo++~\cite{abdullah2024ego2exo,abdullah2026egoexopp} demonstrate the utility of augmented third-person visuals to increase situational awareness during underwater cave exploration; see Fig.~\ref{fig:haptics_xr}\,(b). In addition to rendering the visuals, an HMD is capable of actively controlling the ROV~\cite{candeloro2015hmd}, freeing the operator's hands for other navigational or manipulation tasks; see Fig.~\ref{fig:haptics_xr}\,(a). Windowed or stereoscopic VR~\cite{hine1994application} further improves perception, when compared against monoscopic VR or desktop-based visuals~\cite{elor2021catching}. Other VR-based frameworks such as TWINBOT~\cite{de2020preliminary} offer virtual teleoperation, which is particularly useful for pilot training. The impact of training and XR visuals is supported by studies showing a reduction in task completion time by over 50\%, even when the control method remains unchanged~\cite{elor2021catching}. 
Simulators and digital twins for XR systems are further discussed in Sec.~\ref{sec:simulator}.

\begin{figure*}[t]
    \centering
    \includegraphics[width=\linewidth]{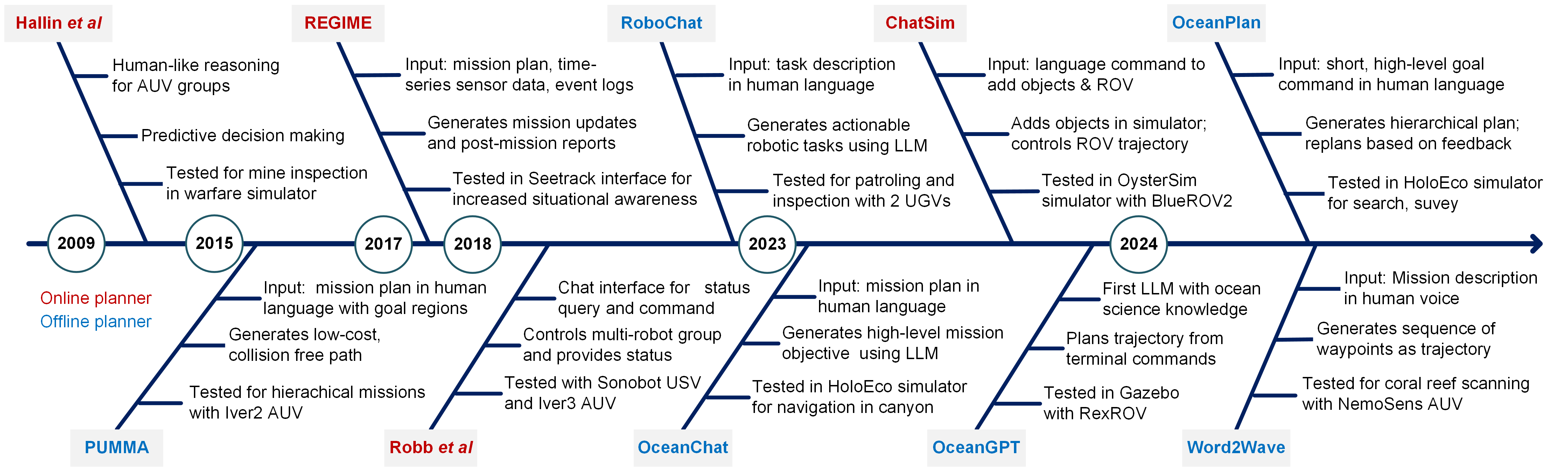}
    \vspace{-2mm}
    \caption{\rebuttal{Progression of natural language functionalities} in subsea telerobotic interfaces and mission management tasks over the past $15$ years is shown. The key works referenced are: Hallin~\etal~\cite{hallin2009using}, PUMMA~\cite{mcmahon2016mission}, REGIME~\cite{hastie2017talking}, Robb~\etal~\cite{robb2018natural}, RoboChat~\cite{li2023robochat}, OceanChat~\cite{yang2023oceanchat}, ChatSim~\cite{palnitkar2023chatsim}, OceanGPT~\cite{bi2023oceangpt}, OceanPlan~\cite{yang2024oceanplan}, and Word2Wave~\cite{chen2024word2wave}.
    %\reviewer{(\#8) figure 4 does not clearly show state of NL integration (it is more, again, pointing on natural language functionalities) and this features cloud be included in table I. Please consider revision of captions.} \Adnan{In the letter to editor, I can explain why these features don't go with Table-I. Table-I compares operational interface products. But this figure summarizes academic literature on NL functions which are not integrated in any product yet.}
    }
    \label{fig:language_timeline}
    \vspace{-4mm}
\end{figure*}

\subsection{Multimodal Interfaces}\label{subsec:haptics}
Besides vision, human sensorimotor control relies on multimodal sensory feedback such as auditory and somatosensory (tactile and proprioceptive) cues to interpret the consequences of initiated actions~\cite{wood2013impact}. Haptic feedback is the most recognized one in enhancing XR systems; researchers have explored innovative methods for simulating haptotactile stimulation (\eg, vibrations and force feedback) synchronized with real-time events~\cite{tian2021haptic}. Haptic-integrated interfaces typically feature a visual feed via a desktop or an HMD and one/multiple haptic feedback devices such as wearable gloves, exoskeletons, body suits, \etc. Preferred choices for haptic feedback include vibrotactile, skin indent, pressure, \etc on finger and wrist regions, as these options only require cheap hardware compared to exoskeletons or full-body haptic suits. \finalrev{To \textbf{clusterize} haptic feedback mechanisms, we define four criteria used in Table~\ref{tab:haptic_interfaces}.}
\begin{itemize}[left=0pt,noitemsep]
    \item \textbf{Feedback location and type} represents the anatomical region and sensory modality used to provide physical stimuli to the operator. While multiple locations and feedback types can be integrated in a single system, the classification is based on the dominant feedback point and function. 
    \item \textbf{Control mechanism} captures the physical console and sensor suites responsible for interpreting and delivering the operator's command/intention to the machine.
    \item \textbf{Ideal use cases} are set by reviewing the user studies and demonstrations reported in the literature; metrics such as task relevance and accuracy improvement are considered for this purpose. Note that the examples are not exclusive, but rather illustrative of commonly reported subsea applications.
    \item  \textbf{Ergonomics} reflects a qualitative estimation of the interface’s ease of use and fatigue factors. This is judged based on user studies (where available) and author-reviewed qualitative descriptions from cited works. 
    %``High'' comfort indicates lightweight, non-intrusive wearables with intuitive mapping (\eg, gloves) while ``low'' refers to bulky, or restrictive setups (\eg, full-body exosuits).
\end{itemize}
\finalrev{Since the latter two criteria require author interpretation, they are explicitly marked and should not be treated as measured performance across all systems.}

The benefits of haptic feedback have been extensively studied for ROV driving and manipulation tasks~\cite{kampmann2015towards}. Ryden~\etal~\cite{ryden2013advanced} present a framework for assisted valve manipulation using real-time haptic feedback to the operator's hand. It guides the robotic arm to stay away from forbidden regions and pulls toward the desired orientation using force feedback. 
% Such force feedback is crucial for the operator's sense of virtual presence and is driven by different sources such as artificial muscle~\cite{sun2009design}, magneto-rheological fluid~\cite{ma2010dynamic,ishizuka2014characterization}, and dc motors~\cite{zhang2017force}. 
To incorporate other external sensory feedback, Amemiya~\etal~\cite{amemiya2009directional} develop a system that combines water pressure and torsion forces to enhance kinesthetic perception. The haptic interface developed by Xia~\etal~\cite{xia2023sensory,xia2023rov} includes the feedback of hydrodynamic state as a vector field, helping the operator to maintain orientation and heading of the ROV in turbulent water.

\begin{figure}[t]
\centering
\vspace{-2mm}
\includegraphics[width=\linewidth]{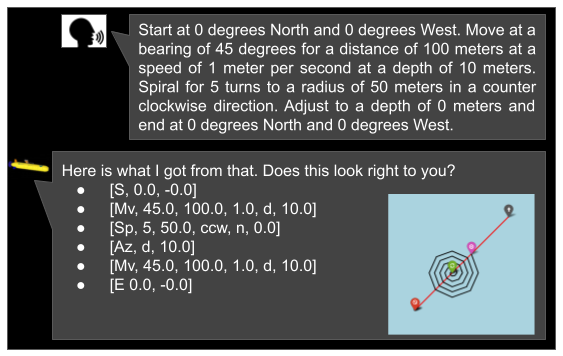}%
\vspace{-2mm}
\caption{An example of a language-guided mission planning interface is shown~\cite{chen2024word2wave}; verbal instructions are parsed into sub-tasks such as {\tt S (start)}, {\tt Mv (move)}, {\tt Sp (spiral)}, {\tt Az (adjust)}, {\tt E (end)} -- to generate mission waypoints.}%
\label{fig:w2w_example}
\vspace{-4mm}
\end{figure}

Whole body motion mapping frameworks~\cite{ha2015whole,xu2021vr} take advantage of body dexterity and movements to control multiple motions of an ROV. However, as mentioned in Table~\ref{tab:haptic_interfaces}, haptic feedback on multiple sensory organs may reduce user comfort~\cite{chen2007human}. To address this, researchers include vestibular feedback using an actuated chair that mitigates HMD-induced motion sickness during long-term operations~\cite{ha2015whole}. Additionally, guiding systems are developed for haptic-supported manipulation that evaluate discomfort in human hand using the Rapid Upper Limb Assessment (RULA)~\cite{mcatamney1993rula}, and direct the hand toward a more comfortable posture~\cite{rahal2020caring}. 
% \JI{Adnan: any other class of work? Seems like an incomplete flow of paragraph} 

\finalrev{Early haptic systems primarily derived feedback from contact-force or camera-based proximity measurements during local manipulation~\cite{ryden2013advanced,kampmann2015towards}. More recent interfaces attempt to convey ambient hydrodynamic state through current or flow sensing and modeled force fields~\cite{xia2023sensory,xia2023rov}. This expansion from contact feedback to hydrodynamic awareness exposes the remaining bottlenecks of reliable underwater sensing, real-time load estimation, and stable perceptual mapping, which motivate the sensing-estimation-rendering pipeline and open problems discussed in Sec.~\ref{subsec:hydro_haptic}.}
% Overall, despite current limitations in haptic devices and ergonomics, visuo-haptic feedback has demonstrated significant potential to enhance the intuitive sense of telepresence.

% Moved to discussion
% Developing intuitive haptic sensation for marine telerobotics remains an ongoing research problem, primarily due to limited communication bandwidth, complex force mechanism, and the unpredictable nature of turbulent waterbodies. For instance, delivering vibrotactile feedback to warn potential collisions, a widely used technique in ground and aerial robotics, proves particularly difficult in water medium due to the restricted capabilities of proximity sensors. Additionally, replicating the sensation of water flow is hindered by the complexities of fluid dynamics and the current limitations of haptic body suits. However, in the long term, haptics could significantly enhance telepresence with novel sensing mechanisms and advanced modeling of hydrostatic and hydrodynamic forces.

\subsection{Integration of Natural Language}

% Natural language and voice commands offer a more intuitive way for operators to interact with complex telerobotic systems. 
% NL-based interfaces leverage the natural human way of communicating and reasoning for more intuitive mission planning and control. The process began as experimental research almost two decades ago~\cite{skubic2004spatial,hallin2009using}, aiming to simplify human-robot communication by enabling operators to issue high-level commands using everyday speech rather than complex control inputs. Classic visual language systems such as RoboChatTag~\cite{dudek2007visual} and RoboChatGest~\cite{xu2008natural} use sequences of symbolic patterns to convey simple instructions from diver to robot/operator, utilizing AR-Tag markers and hand gestures, respectively. Recently, natural language processing (NLP) with deep learning networks has allowed the smooth conversion of spoken/written language into actionable commands for robotic systems~\cite{li2023robochat}. Early NL interfaces focus on basic navigation commands or status queries such as ``take a left", ``report battery status"~\cite{matuszek2013learning}, and have since evolved to support contextual understanding. Fig.~\ref{fig:language_timeline} provides a chronological overview of NL functionalities in telerobotic control and mission management.

\finalrev{NL interfaces translate everyday language into mission or control commands. Early studies explored spoken spatial instructions~\cite{skubic2004spatial,hallin2009using}, while visual-language systems such as RoboChatTag and RoboChatGest encoded diver commands through markers or gestures~\cite{dudek2007visual,xu2008natural}. More recent NLP-based systems map spoken or written instructions to actions~\cite{li2023robochat}, progressing from simple navigation and status queries~\cite{matuszek2013learning} toward contextual mission specification, as summarized in Fig.~\ref{fig:language_timeline}.}

\finalrev{Contemporary mission planners, including OceanChat, Word2Wave, OceanPlan, and AquaChat, convert high-level descriptions into structured actions and trajectories~\cite{yang2023oceanchat,chen2024word2wave,yang2024oceanplan,akram2025aquachat}. Typically, the interface parses the instruction, resolves its semantic content, and maps it to executable mission primitives~\cite{mcmahon2016mission}. Fig.~\ref{fig:w2w_example} illustrates this process for Word2Wave without requiring the operator to specify each low-level waypoint. NL interaction has also been extended to target specification for autonomous underwater manipulation~\cite{lopez2026enhanced}.}

\finalrev{Online interfaces instead maintain dialogue during operation. REGIME summarizes sensor data and mission events on request~\cite{hastie2017talking}, while the interface of Robb~\etal~allows an operator to query and coordinate multiple robots through a designated leader~\cite{robb2018natural}. ChatSim uses dialogue to configure virtual environments and issue navigation objectives~\cite{palnitkar2023chatsim}. Such interaction provides flexibility but is vulnerable to dialect, pronunciation, and linguistic variation~\cite{fleisig2024linguistic}; offline planning permits commands to be reviewed before execution, whereas online systems require explicit ambiguity detection and confirmation. These challenges are discussed in Sec.~\ref{subsec:NL_challenges}.}

\finalrev{Across modalities, no interface is universally preferable. Visual and XR feedback improve spatial awareness, haptics supports contact-rich manipulation, and NL interaction is better suited to mission specification and supervisory control~\cite{elor2021catching,xia2023rov,martin2020application,yang2024oceanplan}. Their suitability depends on task complexity, environmental conditions, communication quality, operator expertise, and autonomy level. Overall, the evidence base remains uneven: many studies use task-specific simulations or limited user studies, while standardized field metrics are rarely reported. We therefore compare reported capabilities and operational roles without claiming a direct performance ranking across the HMIs. For more comprehensive quantitative comparison, future evaluations should include task-completion time, error and recovery rates, operator workload, communication conditions, autonomy level, and operator expertise.}

% The current challenges with integrating language models into telerobotic systems are multifaceted. First, the computational demands of LLMs may exceed the capabilities of resource-constrained underwater systems. To address this, researchers are developing smaller language models (SLMs) while technologies such as LoRA~\cite{hu2022lora} are helping to adapt LLMs for real-time applications in edge devices. Second, ensuring that the robot accurately interprets commands and understands spatial references requires sophisticated models, extensive training data, and robust error-handling mechanisms~\cite{li2023robochat}. The consequences of misinterpretation can be severe in high-risk subsea operations, hence current research focuses on making these systems more resilient, context-aware, and capable of handling complex, multi-step instructions~\cite{valluri2024integrating}. Moreover, online commands can be error-prone due to variations in dialect, tone, or pronunciation~\cite{fleisig2024linguistic}. Instead, an offline planning interface allows the operator to review and correct for potential errors before launch. Further discussions on this are presented in Sec.~\ref{sec:discussion}.
% In the near future, we anticipate a significant rise in language-driven teleoperation interfaces, especially for marine robots.

\section{Simulators and Digital Twins}\label{sec:simulator}
Simulators offer a valuable alternative for prototyping, error debugging, and pilot training for marine robotic systems in a controlled, risk-free setting, without requiring expensive logistics and access to real waterbodies. 
\finalrev{However, they are task-dependent support tools rather than required components of teleoperation. Conventional simulators support offline prototyping and algorithm development, while vehicle models can bridge short feedback gaps. Additionally, digital twins are most relevant during live bidirectional synchronization and support online monitoring, fault diagnosis, or adaptive planning~\cite{jones2020characterising,van2023digital}.}
% These virtual solutions allow researchers to evaluate and refine underwater robotic systems in .
% Researchers explore innovative ways to leverage computer-based virtual solutions \ie simulators, that provide rich, immersive environments for testing and refining underwater robotic systems. 
% They not only replicate the environmental dynamics but also support the integration of digital replicas of physical systems, facilitating the development of control algorithms and mission planning strategies.
% , facilitating the testing and development of robust control algorithms and mission planning strategies for real robots. Fig.~\ref{fig:tank_chase} shows an example of underwater grasping with a digital twin rendered in the HoloOcean simulator. 

\subsection{Digital Twin (DT)}
% \reviewer{(\#9) The Digital Twin concept is misleading as it, according to the description, more in terms of shadowing than as a DT. Please, enlarge/improve the section.}
A digital twin refers to a simulated replica of a real robot that possesses similar attributes and can mimic the motion of its real-world pair within a virtual environment~\cite{jones2020characterising}. The concept of DT for field robots was first presented by NASA, where the digital entity was defined as an integrated multi-physics, multi-scale, probabilistic simulation of a vehicle or system that mirrors the life of its real-world peer~\cite{shafto2010draft}. This project envisioned four applications of the DT technology including pre-mission simulation, real-time modification, monitoring, and onboard forensics for comprehensive analyses.

% \begin{itemize}[left=0pt]
%     \item Pre-mission simulation: The twin will simulate the vehicle's future missions before launch, enabling studies of mission parameters, anomalies, and mitigation strategies to enhance mission success.

%     \item Real-time flight mirroring: The twin will continuously update and mirror the actual flight with real-time sensor data, predicting flight status for ongoing analysis and adjustment.
    
%     \item In-situ forensics: The twin will allow diagnosis and analysis of potential catastrophic faults in real-time and address anomalies during the flight.
    
%     \item Mission modification analysis: The twin will assess the impact of unforeseen mission parameter changes (e.g., failed actuators), and analyze new load distributions to inform mission decisions.

\begin{figure}[t]
    \centering
    \includegraphics[width=\linewidth]{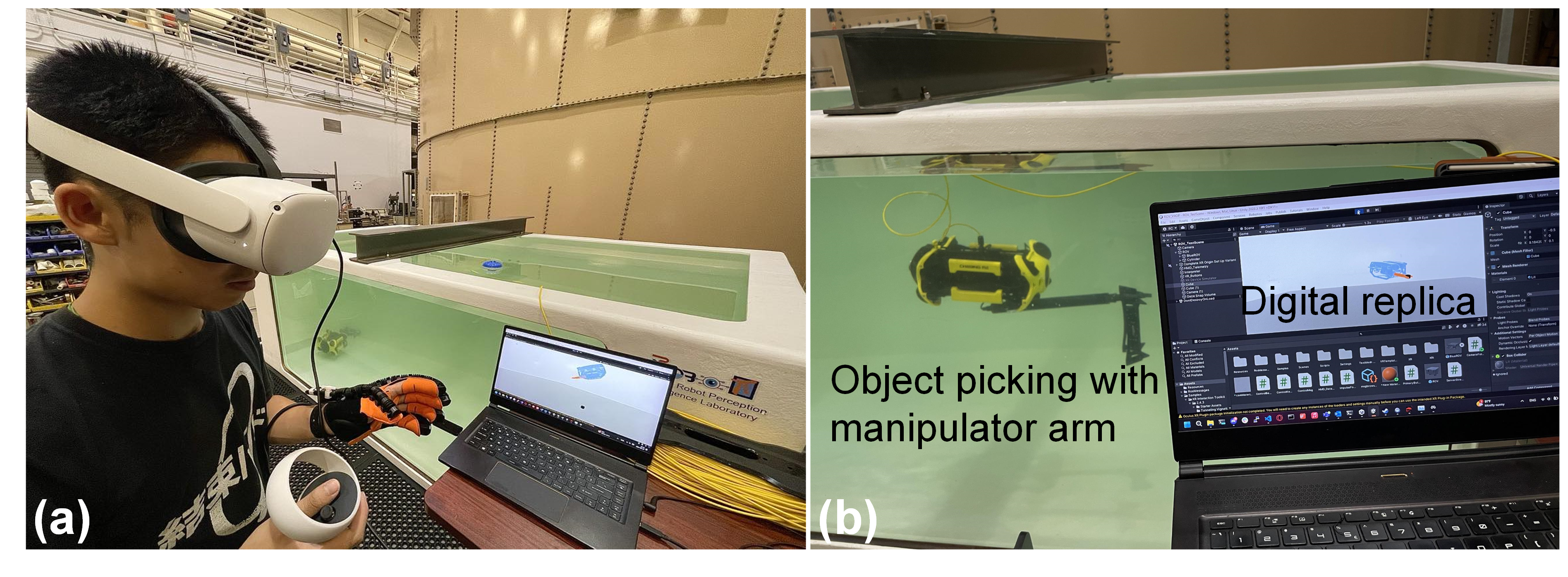}%
    \vspace{-2mm}
    \caption{A \rebuttal{digital shadow} of an underwater ROV is shown: (a) a human is teleoperating it for object grasping; (b) the replica rendered in HoloOcean~\cite{potokar2022holoocean} simulates the action in the virtual environment.
    }
    \label{fig:tank_chase}
    \vspace{-3mm}
\end{figure}

% \end{itemize}
\rebuttal{Although the terms digital \textit{shadow} and \textit{twin} are used interchangeably in literature, there is a clear distinction between them. A shadow is a passive digital replica that mirrors the real-time entity without active predictive modeling or feedback loops~\cite{sepasgozar2021differentiating}. It typically supports one-way (\ie, real2sim) data flow and is limited to basic monitoring or visualization. A true DT would include high-fidelity representations of vehicle dynamics, sensor fusion models, environmental interactions, and a bidirectional data pipeline enabling predictive simulation~\cite{sjarov2020digital,yu2026sixdimensional}. 
% DTs exhibit varying levels of complexity, ranging from basic motion replication to comprehensive emulation of physical and operational attributes. 
For instance, the simplified setup shown in Fig.~\ref{fig:tank_chase} does not offer predictive simulation; however, the digital robot mimics the motion of the real ROV, thereby qualifying as its shadow rather than a twin.} 
% For more sophisticated DTs, the digital replica must mirror key physical properties (\eg dimensions, weight, buoyancy), energy consumption profile, sensor configurations, and interaction with the external environment.
% Examples of some SOTA simulators and their shadow/twin collections include BlueROV2 in UNav-Sim~\cite{amer2023unav}, Triton XLX in PaleBlue~\cite{paleblue} and VMAX simulator~\cite{vmax}, VideoRay Pro5 and Defender in Era's VideoRay simulator~\cite{era_videoray}, and Girona500 in UWSim~\cite{prats2012open}.

\begin{table*}[t]
    \centering
    \caption{A comparison of underwater simulators is shown. Basic state estimation sensor plugins (\eg, IMU, GPS, camera, depth) are common for all simulators; therefore omitted here; $^\diamond$: DT of a real ROV; other notations are explained in Sec.~\ref{sec:scope}.
    % \reviewer{(\#10) About table III, I expected to read/suggest a column about optimal usage.}
    }
    \resizebox{\textwidth}{!}{
    \begin{tabular}{lllll}

    \vspace{-9mm} \\
    \multicolumn{5}{l}{{\small (a) Comparison based on sensor integration capability, scenario libraries, and available robotic agents.}} \\
      \Xhline{2\arrayrulewidth}
       \rowcolor[rgb]{0.9,0.9,0.9}  &  Sensor Plugin$^\dagger$$^\ddagger$ & Built-in Scenarios$^\dagger$ & Robot Models$^\dagger$$^\ddagger$ & \rebuttal{Optimal Usage$^\blacklozenge$} \\
      \Xhline{2\arrayrulewidth}
      PaleBlue~\cite{paleblue} & Laser & Reefs, oil and gas rigs & Triton XLX\,$^\diamond$ & VR-supported rig inspection\\
      Abyssal~\cite{abyssal} & - & Rigs, sand cloud, pipelines & Abyssal ROV & Manipulator control \\
      M8-R Sim~\cite{m8rsim} & - & 140+ coral reefs with starfish & M8-R ROV & Education \& novice training \\
      Era VideoRay~\cite{era_videoray} & Sonar, lights & 6 scenarios with shipwrecks  & VideoRay Pro5\,$^\diamond$, Defender\,$^\diamond$ & Training for VideoRay ROVs\\
      GRi VROV~\cite{gri} & Sonar & Pipelines and rigs  & LittleGeek, BigGeek & Maintenance, rescue, mine detection\\
      UUV Simulator~\cite{manhaes2016uuv} & Sonar, DVL, CPC & Shipwrecks, coast, BOP panels & A9\,$^\diamond$, LAUV\,$^\diamond$, Saga\,$^\diamond$, RexROV & Perception \& control study\\
      UUVSim~\cite{zhang2024uuvsim} & Sonar & Open water, current, seabed & 6T UUV, RexROV & Hydrodynamic study, DL training\\
      HoloOcean~\cite{potokar2022holoocean} & Sonar, beacon, DVL & Dams, pier harbors, pipelines & 6 ROV agents & Sim2Real, multi-agent control\\ 
      UWSim~\cite{prats2012open} & DVL, multi-beam sonar  & Dredging, pipelines, amphora  & Girona500\,$^\diamond$ & Planning, control, HIL study\\ 
      UNav-Sim~\cite{amer2023unav} & LiDAR & Algae, sandy pipelines & BlueROV2 heavy\,$^\diamond$ & Navigation \& inspection study\\ 
      MARUS~\cite{lonvcar2022marus} & LiDAR, sonar, DVL, AIS & Ships, human divers, seabed & One AUV & Acoustic sensing study\\ 
      StoneFish~\cite{cieslak2019stonefish} & Sonar, DVL & Sea surface, seabed, wind, sun & Girona500\,$^\diamond$ & Localization \& navigation study\\ 
      NaviSuite~\cite{navisuite} & Sonar, DVL, USBL & Cables, pipelines, and ship anchors & VideoRay Defender\,$^\diamond$ & Inspection, intervention \\ 
      VMAX Simulator~\cite{vmax} & Sonar & Cables, ship with crane & Triton\,$^\diamond$, Comanche\,$^\diamond$, Mojave\,$^\diamond$ & Manipulator control\\ 
      ACSM~\cite{acsm} & - & Rigs and pipelines & Triton XLX\,$^\diamond$ & Deployment \& manipulator control\\
      Unicom~\cite{unicom} & - & Rigs with valves, knobs, doors  & 2 work-class ROVs & Maneuvering risk assessment\\ 
      \bottomrule
    \end{tabular}   
    }
    \resizebox{\textwidth}{!}{
    \begin{tabular}{lccccccccccccccc}
    \vspace{0.5mm} \\
    \multicolumn{16}{l}{{\normalsize (b) Comparison based on the availability of various features and supported platforms.} 
    %Note that, interacting with dynamic obstacle is critical
    } \\
    %\multicolumn{13}{l}{ for safe teleoperation, hence highlighted in this comparison.} \\
    \Xhline{2\arrayrulewidth}
      &  Pale & Abyssal & M8-R & Era & GRi  & UUV  & Holo & UW  & Unav & MARUS & Stone & Navi & VMAX & ACSM & Unicom \\
     &  Blue & Sim & Sim & VideoRay & VROV & Simulator & Ocean & Sim & Sim &  & Fish & Suite & Simulator &  & \\
     %&  \cite{paleblue} & \cite{abyssal} & \cite{m8rsim} & \cite{era_videoray} & \cite{gri} & \cite{zhang2024uuvsim} & \cite{potokar2022holoocean} & \cite{uwsim} & \cite{amer2023unav} & \cite{navisuite} & \cite{acsm} & \cite{unicom} \\
     
    \Xhline{2\arrayrulewidth}
    \cellcolor[rgb]{0.9,0.9,0.9}{Open Source} & \cross & \cross & \tick & \cross & \cross & \tick & \tick & \tick & \tick & \tick & \tick & \cross & \cross & \cross & \cross \\
    \cellcolor[rgb]{0.9,0.9,0.9}{Haptic Support$^\ddagger$} & \tick & \tick & \cross & \cross  & \tick & \cross & \cross & \cross & \cross & \cross & \cross & \cross & \cross & \cross & \cross\\
    \cellcolor[rgb]{0.9,0.9,0.9}{ROS Support} & \tick & \cross & \cross & \cross & \cross & \tick & \tick & \tick & \tick & \tick & \tick & \cross & \cross & \cross & \cross \\
    \cellcolor[rgb]{0.9,0.9,0.9}{Training Modes} & \tick & \tick & \tick & \tick & \tick & \cross & \cross & \cross & \cross & \cross & \cross & \tick & \tick & \tick & \tick \\
    \cellcolor[rgb]{0.9,0.9,0.9}{Dynamic Obstacles} & \tick & \cross & \tick & \cross & \tick & \cross & \cross & \tick & \tick & \cross & \cross & \cross & \tick & \cross & \cross \\
    %\midrule
    \cellcolor[rgb]{0.9,0.9,0.9}{OS (W,M,L)}  & \tick\cross\cross & \tick\cross\cross  & \tick\cross\tick & \tick\tick\tick & \tick\cross\cross & \tick\tick\tick & \tick\cross\tick & \tick\tick\tick & \tick\cross\tick & \tick\cross\tick & \tick\tick\tick & \tick\cross\cross & \tick\cross\cross &  \tick\cross\cross & \tick\tick\tick \\
    
    \bottomrule
    \end{tabular}
    }
    \vspace{-3mm}
    \label{tab:simulator}
\end{table*}

The use of DTs has been reported in different branches of robotics, including aerial, manufacturing, surgical, space, and marine robots. 
% All DT systems are built upon some common backends with domain-specific techniques for data acquisition and rendering~\cite{jones2020characterising}. 
Generally, the raw sensor data acquired from the robot is processed with physics engines and reconstructed as a high-fidelity DT model~\cite{jones2020characterising}. Considering the complex nature of subsea environments, Van~\etal~\cite{van2023digital} identify the communication between the two entities as a fundamental component of a successful subsea DT system. The authors highlight three use cases for DT: \textbf{(i)} ROV launch and recovery training, \textbf{(ii)} failure prediction and remaining useful life estimation, and \textbf{(iii)} trust-aware DT development to reduce operator's decision-making time in critical situations. However, other literature~\cite{yang2022digital,zuba2022pairing} argues that DT systems do not require a fully reliable or accelerated communication mode. Instead, during a communication disruption, a predictive DT system can be utilized to create a realistic representation as if communication had remained uninterrupted. 
% DT systems have also been deployed in underwater sensor networks to test communication reliability and synchronization accuracy between multiple nodes~\cite{barbie2021developing}. 
More recently, Adetunji~\etal~\cite{adetunji2024digital} integrate SLAM and path planning capabilities in a DT system. Other applications of marine DT include inspection~\cite{kleiser2020towards}, assistive control~\cite{xia2023sensory}, trajectory planning~\cite{yang2022digital}, \etc.

% M8-R Sim and Unicom is web-based

\subsection{Underwater Simulators}
A simulator is essential to work with DTs as well as to mimic the external environment parameters. However, given the complex fluid dynamics and other physicochemical variations, it is extremely difficult to design an accurate physics-based model of an underwater environment. Engineers have created aquatic environments with physics and game engines (\eg Unity, Blender, WebGL3D, ODE) as backends and have integrated basic properties of the medium such as buoyancy, waves, current profile, \etc; examples include UUV Simulator~\cite{manhaes2016uuv} and UWSim~\cite{prats2012open} that support the Gazebo plugin~\cite{koenig2004design} as well. Independently designed virtual ROVs or DTs of real ROVs are rendered in these platforms to test their kinematic properties and interactions with the surroundings. Researchers have studied different underwater vehicles such as gliders~\cite{yang2022digital,hu2023construction}, ROVs~\cite{zuba2022pairing,manhaes2016uuv}, AUVs~\cite{kleiser2020towards}, and humanoid robots~\cite{khatib2016ocean} in these simulators. Yang~\etal~\cite{yang2022digital} investigate a glider trajectory in a deep virtual ocean. Their simulations show the effects of the current gradient across the water column on the glider's trajectory, especially during the vertical dives. Hu~\etal~\cite{hu2023construction} model a blended-wing-body underwater glider and investigate the accuracy of a pitch controller under ocean current influence. 

% \begin{figure}[t]
%     \centering
%     \includegraphics[width=\linewidth]{images/simulators3.png}%
%     \vspace{-1mm}
%     \caption{Snapshots of a few marine robotic simulators are shown; the top row shows proprietary simulators for professional use~\cite{paleblue,abyssal,vmax}; the bottom row shows open source simulators mainly used for academic research~\cite{manhaes2016uuv,potokar2022holoocean,prats2012open}.}
%     \label{fig:simulators}
%     \vspace{-3mm}
% \end{figure}

\finalrev{Table~\ref{tab:simulator} reveals two broad roles of simulators and digital replicas. Proprietary platforms target operator training and industrial workflows, offering large-scale infrastructure models, vehicle-specific twins, and task-oriented scenarios with limited extensibility. Open-source platforms generally provide fewer training scenarios but greater access to physics, sensor, ROS, and robot models, making them more suitable for research and algorithm development. Therefore, simulator suitability depends less on feature count than on the intended objective: operational training favors scenario and vehicle fidelity, whereas algorithm development favors configurability, sensor diversity, and software integration.}

\section{Human-Machine Shared Autonomy} \label{sec:autonomy}
%\Adnan{How about ``Human-Machine Intelligence Sharing and Interaction"?}

\finalrev{Subsea telerobotic systems span three levels distinguished by machine decision-making and allocation of control authority. In direct teleoperation, the operator issues low-level commands and depends on timely feedback; delay and packet loss therefore affect stability, transparency, and workload~\cite{zhang2024adaptive,mitra2008model}. In shared autonomy, the operator sets goals or supervises while the robot performs functions such as stabilization, obstacle avoidance, or planning. This reduces continuous input but requires command arbitration, authority transitions, and clear mode awareness~\cite{lawrance2019shared,xia2024human}. Autonomous systems execute planned missions with minimal intervention, shifting the HMI toward mission specification, status monitoring, fault alerts, and abort or recovery commands~\cite{yan2024teleoperation}. Increasing autonomy can reduce continuous communication dependence, but increases the need for predictable behavior and safe handover.}
% Subsea robotic systems are categorized into three types according to their intelligent decision-making capabilities: \textbf{(i)} direct teleoperator control with no intelligence, \textbf{(ii)} shared/collaborative reasoning, and \textbf{(iii)} autonomous/self-sufficient systems. In direct teleoperation, the operator has total control over the robot's motion and actions, generally through a live feed and a motion controller. However, these methods are constrained by communication latency and operator workload, making them less efficient for long-term complex tasks. Shared autonomy features bridge the gap between the human and the machine by allowing robots to perform certain tasks autonomously while the operator manages high-level decisions~\cite{lawrance2019shared,xia2024human}. This reduces the operator's cognitive load, compensates for communication delays, and enhances operation safety. In contrast, AUVs function independently with minimal human intervention~\cite{yan2024teleoperation}. These robots rely on advanced algorithms for perception, decision-making, and execution, making them ideal for repetitive tasks in familiar environments. However, their utility is constrained by environmental uncertainties, the cost of logistics, and the limitations of current AI-based technologies~\cite{barker2020scientific}. 
% \JI{again and "often" statement with no justification. What is often? you mean sometimes? or some of the features or some of the robots? Where are citations to back your claim??????}

\begin{table*}[t]
    \centering
    \caption{The table summarizes various types of intelligent machines and their autonomous task-sharing capabilities reported across the subsea telerobotics literature. \rebuttal{Criteria are defined in Sec.~\ref{subsec:shared_autonomy}, notations are explained in Sec.~\ref{sec:scope}}.
    % \reviewer{(\#11) Unclear is, in my opinion, the definition of autonomy categories and intelligence type of table IV. The table needs a clear descriptive support for the sake of consistency.}
    }
    \vspace{-1mm}
    \resizebox{\textwidth}{!}{
    \begin{tabular}{l||ll|ll|ll}
    \Xhline{2\arrayrulewidth}
    \rowcolor[rgb]{0.9,0.9,0.9} Type of & \multicolumn{2}{c|}{\underline{Human (Teleoperator) Side}} &  \multicolumn{2}{c|}{\underline{Machine (Underwater ROV) Side}}    & Selected \\ 
    \rowcolor[rgb]{0.9,0.9,0.9} Intelligence & Responsibility$^\dagger$$^\blacklozenge$ & Cognitive Load$^\blacklozenge$ & Intelligent Capabilities$^\dagger$ & Major Applications$^\blacklozenge$  & References \\ 
    \Xhline{2\arrayrulewidth}
     Command  & Issue high-level  & Medium & Translate to low-level action & Surveying, grasping,  & \cite{di2016advanced,senft2021task} \\
     interpretation & commands & & and check feasibility & inspections, interventions  & \\ \hline
     Online course & Set new goal point & Low & Update local planner and & Dynamic path planning  & \cite{lee2016development,yang2024oceanplan} \\ 
     adjustment  & or trajectory &  &  mission parameters & in uncertain environment &  \\  \hline
     Error & Supervise and act  & Medium & Diagnose issues, report,  & Fault tolerant system for  & \cite{kaeli2016real,xiang2017intelligent} \\ 
     handling & upon anomaly & & attempt resolution & long-term exploration &  \\  \hline
     Multi-robot & Operate the  & Medium & Coordinate and control  & Coverage, inspection,  & \cite{matos2016multiple,howard2006experiments} \\ 
     collaboration & leader ROV &  & follower ROVs & search and rescue  & \cite{patel2024multi} \\  \hline
     Mission  & Set mission & Low & Generate waypoints,  & Exploration, mapping,  & \cite{yang2023oceanchat,yang2024oceanplan} \\ 
     planning & objectives &  & execute motion  &  area coverage  & \cite{yang2022digital} \\  \hline
     Learning from  & Demonstrate & Initially high & Learn knowledge base,  & Routine maintenance,  & \cite{havoutis2019learning,maurelli2016pandora} \\ 
     demonstration & relevant tasks &  & adapt to novel tasks  & sample collection  & \cite{somers2016human} \\ \hline
     Context-aware  & Provide contextual & Medium & reason a context, & Inspection, surveillance,  & \cite{jacobi2015autonomous,ruscio2023autonomous} \\ 
     exploration & cues or guidance &  &  plan accordingly & environmental monitoring  &  \\ 
     % \hline
     % Docking & Guide the robot  & Low & Identify docking point, &  Charging, data transfer,   & \cite{stokey2001enabling,trslic2020vision} \\
     % assistance & close to the dock &   &  self-correct fine motion  &  repair, recovery   &  \\
   
    \bottomrule
    \end{tabular}
    }
    \vspace{-3mm}
    \label{tab:autonomy}
\end{table*}

\subsection{Direct Teleoperation Mode}
% \JI{are we talking about "control?" or direct teleopration as a "mode"?}
Direct or hands-on teleoperation involves controlling the robot using low-level commands without intelligent decision-making by the robot. It allows for responsive control, enabling precise adjustments that automated systems may struggle with. Despite high cognitive demand, it remains the preferred approach for manipulation tasks involving high risk and fine precision, where the operator controls one joint angle at a time in a \textit{joint-by-joint} fashion~\cite{jamieson2013deep}; examples include inspection~\cite{ruscio2023autonomous}, valve control~\cite{zhang2024adaptive}, and benthic sampling~\cite{jamieson2013deep}.

\finalrev{Communication delay and packet loss can destabilize bilateral control and distort the force or motion feedback perceived by the operator. Adaptive bilateral controllers preserve stability under uncertain delay, although conservative stabilization may reduce transparency~\cite{anderson1991stable,zhang2024adaptive}. Predictive displays and model-mediated teleoperation instead estimate remote behavior locally to improve perceived responsiveness, but their reliability depends on model accuracy and explicit communication of prediction uncertainty~\cite{mitra2008model,rank2016predictive}. For longer delays or link degradation, supervisory, shared, and local autonomy can maintain low-level stabilization, collision avoidance, or task execution without continuous commands~\cite{gancet2015dexrov}. These autonomy transitions require safe command arbitration and clear indication of the active control mode.}

% These systems are challenged by the inherent communication delays in underwater environments, which can reduce control precision and responsiveness. To address this, researchers have developed adaptive control strategies to compensate for both model uncertainties and external disturbances during teleoperation~\cite{zhang2024adaptive}. Furthermore, electromagnetic and optical communication modes offer promising solutions for establishing high-speed reliable links~\cite{talebi2024blueme}, improving the real-time commands and feedback flow in direct teleoperation systems.

\subsection{Human-Robot Collaborative Reasoning}\label{subsec:shared_autonomy}
The goal of collaborative decision-making systems is to offer a blend of human oversight along with automated functions~\cite{li2023proactive}. \finalrev{These systems allow operators to set high-level objectives while the robot handles low-level operations, reducing cognitive load and operator effort~\cite{xia2024human,tu2025unified}.} For instance, an operator can issue \textit{hovering} command; the robot determines its state using an IMU, DVL, and depth sensors to correct for any drift motion to remain stationary. More advanced reasonings are achieved in a supervisory manner (for a single robot) or with a leader-follower strategy (for a group of robots). 
% Table~\ref{tab:autonomy} provides an overview of the SOTA intelligent capabilities of underwater robots and their task-sharing mechanisms with humans. 
\rebuttal{Table~\ref{tab:autonomy} aims to present a structured overview of current capabilities and highlight how autonomy is distributed in SOTA subsea HMI systems. The \textbf{clustering} is motivated by a need to understand and compare the evolving roles of humans and machines in underwater missions. In modern telerobotic systems, autonomy is no longer binary (teleop vs. autonomous) but exists on a spectrum. Thus, the table categorizes task-sharing models by considering: \textbf{(i)} the nature of human intervention, \textbf{(ii)} the degree of machine intelligence, and \textbf{(iii)} the operational context. The cognitive load classification is informed by findings from user studies and qualitative insights reported by other researchers.
% The criteria and definition of intelligence types are set as follows:
% \begin{itemize}[left=0pt,nosep]
%     \item Command interpretation: translate high-level mission goal into low-level control commands.
%     \item Course adjustment: dynamically adapt actions if mission objective is changed.
%     \item Error handling: diagnose and self-correct (if possible) faults or anomalies during operation.
%     \item Multi-robot collaboration: coordinate and serve as the intermediary between the operator and the team.
%     \item Mission planning: generate local plans based on a global objective.
%     \item LfD: adapt strategies to future scenarios based on past demonstrations.
%     \item Context-aware exploration: reasons like a human upon a given context and acts accordingly.
%     \item Docking assistance: corrects fine motion and overrides operator command (if necessary) for safe docking.
% \end{itemize}
}

\finalrev{Adaptive HMI behavior requires context monitoring coupled with explicit authority management. The interface can track mission phase, environmental conditions, communication quality, vehicle health~\cite{saad2026aquachatpp}, and operator-workload indicators. It can then prioritize warnings, suppress nonessential information, invoke local stabilization or replanning, increase autonomous assistance, or request operator takeover. To avoid mode confusion, each transition should communicate its trigger, the current allocation of control authority, and the available override or recovery action~\cite{lawrance2019shared,yang2024oceanplan,kaeli2016real}.}

For instance, mission planning, online mission adjustment, and fault diagnosis systems benefit from shared autonomy while keeping the \textit{operator in the loop}~\cite{teigland2020operator}, but without demanding continuous oversight. Research in supervised planning and navigation demonstrates that a new mission path suggested by a circumstance-aware robot is often safer and reduces task completion time~\cite{senft2021task,yang2024oceanplan,islam2021robot}. The GALILEO planning interface adjusts the course of actions and resource allocations based on high-level goals set by the operator~\cite{jackson2022scheduling}. Additionally, intelligent robots are capable of detecting and reporting internal faults and data anomalies~\cite{kaeli2016real}, reducing the need for constant operator oversight. 
% Such distributed control also allows the operator to focus on mission objectives while offloading low-level tasks to the robot. 
Sharing scientific objectives with a context-aware robot is beneficial too, as the robot can search for specific cues or features under operator guidance. The CView project~\cite{jacobi2015autonomous} provides an example; the operator defines the object of interest (\eg, wall, vessel, sluice) and the features to detect (\eg, cracks), enabling the robot to survey the objects and report key findings.

SUPERMAN, the first supervisory subsea teleoperation interface, was proposed by NASA in $1978$~\cite{sheridan1978human}; however, the development has been slow due to the difficulty of designing high-level autonomy considering the environmental challenges. SAUVIM ($1998$)~\cite{yuh1998design} is one of the earliest semi-autonomous vehicles with a 6-DOF manipulator arm developed for deep water\invis{($6000$\,m)} intervention tasks. More recently, supervisory control frameworks such as DexROV~\cite{gancet2015dexrov} and SHARC~\cite{phung2024shared} have been developed where the operator performs remote manipulation tasks using language commands and hand gestures~\cite{senft2021task}. The GRASPER project~\cite{sanz2013grasper} utilizes a pre-scanned scene point cloud to inform the robot about obstacle locations. The autonomous action is rendered in the UWSim simulator~\cite{prats2012open} in real time, allowing the user to supervise grasping tasks in a virtual world.

Moreover, contemporary researchers have developed adaptive supervision systems where the robot understands the operator's overarching intention and adjusts the low-level reactions accordingly. Lee~\etal~\cite{lee2016development} propose a shared teleoperation method where humans sketch a desired path on the visual console, guiding the intelligent system, which plans an optimum path along that drawn trajectory and executes the motion. The adaptive supervision system built within the GreenSea interface provides operators customized assistance based on their skill levels~\cite{lawrance2019shared}. This helps novice pilots achieve performance comparable to that of experts.

% Supervisory control has been employed for precise manipulation, path-planning, and navigation tasks where the human operator acts as a supervisor and uses his cognitive ability to take high-level control decisions. Those decisions are translated into low-level tasks and performed by the robot on it's own. 
% Such systems improves efficiency and reduce operator's fatigue compared to direct teleoperation. 

% \JI{This part of the writing seems like V1, not properly revised}

For more complex scenarios, learning from/by demonstration (LfD/LbD) represents another collaborative paradigm, where the robot learns a \textit{knowledge base} from operator-provided demonstrations and uses it to autonomously perform analogous tasks in the future. 
\finalrev{It converts operator demonstrations into reusable robot skills, unlike reward-driven RL~\cite{correia2024survey}. Subsea applications include drilling and cutting, valve operation, and hot-stabbing~\cite{havoutis2016learning,carrera2014learning,havoutis2017supervisory}. Its main limitations are sensitivity to demonstration quality and discrepancies between the demonstrated and operational spaces. Probabilistic path planning can accommodate variable demonstrations~\cite{somers2016human}, while hidden semi-Markov task representations have improved transfer in simulated manipulation~\cite{havoutis2019learning}, although vehicle dynamics and environmental disturbances remain insufficiently evaluated. PANDORA further demonstrates that autonomous inspection and intervention require a semantic knowledge base rather than only predefined trajectories~\cite{maurelli2016pandora}.}

\finalrev{In multi-robot systems, an operator may directly control a leader while locally autonomous followers maintain formation and trajectory~\cite{islam2021robot,edwards2004leader}. Such teams extend spatial coverage for search and rescue, infrastructure inspection, mapping, and cooperative localization~\cite{matos2016multiple,patel2024multi,hover2012advanced,howard2006experiments,kim2020cooperative}. Although some formulations place the operator outside the swarm after mission initialization~\cite{zhou2021survey}, others retain direct leader control while followers estimate their relative poses autonomously~\cite{karras2015towards,yao2023image}. Recent work increasingly considers heterogeneous teams rather than identical vehicles~\cite{kalantar2007distributed}, assigning tasks according to vehicle capabilities and operational state~\cite{cai2023cooperative}. For example, a surface vehicle can relay communication between the operator and multiple ROVs during pipeline inspection~\cite{patel2024multi}.}

\subsection{Autonomous Systems}\label{subsec:AUVs}
% \JI{are we talking about AUVs+humans here or still in telerobotics. Both are fine, make it clear}
%\JI{Not clear why are we talking about AUVs suddenly. make the distinction that fully autonomous system do not fall under the umbrella of teleoperation. Then also justify why are you talking about AUVs here?}
% \reviewer{(\#12) In Section 4, telerobotic systems are categorized into three types: (i) direct teleoperator control with no intelligence, (ii) shared/collaborative reasoning, and (iii) autonomous/self-sufficient systems. It is essential for the authors to fully demonstrate the research status, limitations, and development trends of fully autonomous systems.}
As opposed to direct teleoperation and shared autonomy schemes, autonomous systems or AUVs operate independently, requiring no control commands from the operator. Subsequently, HMI for an AUV is primarily designed for status updates and mission management rather than direct control. However, even fully autonomous systems rely on human operators for mission planning, monitoring, and intervention. %This highlights the critical role of AUV interface design in ensuring the safety and effectiveness of autonomous missions.

Generally, the interface includes a software console, a surface control module (SCM), and a transducer to maintain real-time communication with the AUV. Available real-time commands at the operator end are typically limited to starting or aborting the mission, requesting status updates, or intervening during critical moments. Contemporary works have integrated more advanced automated features into these interfaces. For instance, the MARES AUV interface automatically performs pre-mission procedures such as estimating mission duration and path length, testing communications, \etc~\cite{abreu2010automatic}. Kaeli~\etal~\cite{kaeli2016real} develop an anomaly detection framework that enables AUVs to focus on high-resolution scanning and transmit only the anomalous data to the operator, reducing data congestion and improving efficiency. A similar approach is reported for crack detection in subsea pipelines using a camera, laser, and multi-beam echosounder; the AUV detects potential cracks and sends the scanned images to an operator for further review~\cite{jacobi2015autonomous}. The operator then requests close-up detailed examination if necessary, or continues the mission. 
% A detailed review of current AUV technologies is outlined in~\cite{sahoo2019advancements}.

% Currently, the top application segments for AUV/UUVs are military and defense, offshore oil and gas drilling,  oceanography, archaeology, and alternative energy harvesting~\cite{nicholson2008present}. Their operational costs are high due to factors including maintenance, repair, data acquisition, personnel training, and mission-specific logistics. 

% \JI{The operator then does what?}
% 

% AUVs represent the cutting edge of marine technology with their history going back to the $1960$s when Stan Murphy and Bob Francois developed the first SPURV (Self-Propelled Underwater Research Vehicle) in the Applied Physics Laboratory at the University of Washington~\cite{widditsch1973spurv}. AUVs have come a long way since then, but achieving true autonomy remains challenging to this day. As discussed earlier, the robots can operate autonomously to some extent, but more complex long-term autonomous missions demand some operator assistance/supervision. Key challenges for AUVs include detecting and avoiding potential collisions inside confined natural spaces (\eg, caves) or around man-made structures (\eg, pipelines, and shipwrecks), localizing and navigating through feature-deprived scenarios, maintaining communication with the operator, \etc~\cite{paull2013auv}.

\vspace{1mm}
\noindent
\rebuttal{\textbf{Current status \& limitations}. Currently, the top application segments for AUV/UUVs are military and defense, offshore oil and gas drilling,  oceanography, archaeology, and alternative energy harvesting~\cite{nicholson2008present}. Their operational costs are high due to factors including maintenance, repair, data acquisition, personnel training, and mission-specific logistics. Consequently, access to AUVs has been restricted to governments and their agencies, industry leaders, and a handful of academic and research labs. Notable platforms such as Kongsberg Hugin (Germany), SAAB Seaeye Sabertooth (Sweden), and Exail A18 (France) have demonstrated long-range deployments for pre-scripted missions with limited surface intervention. Military-grade AUVs such as L3Harris IVER (US) and Atlas Elektronik SeaCat (Germany) offer surveying, intelligence gathering, mine disposal, \etc. 
% A9-E ECA (France), Teledyne Gavia (US), . 
A lightweight, low-cost category known as micro-AUV is gaining popularity for academic and research purposes; examples include NemoSens (France), Seaber Yuco series (France), and EcoSUB (UK). 
}

\rebuttal{
However, true autonomy, involving decision-making, adaptive planning, and high-level semantic understanding, remains an ongoing research challenge. We summarize the following limitations by reviewing the SOTA AUV literature.
\begin{itemize}[left=0pt,noitemsep]
    \item \underline{Environmental uncertainty:} Underwater environments are highly dynamic, with varying turbidity, current, and visibility, making perception and long-term localization unreliable.
    \item \underline{Communication and energy constraints:} Lack of real-time power and data transfer prevents operator supervision and long-term deployments.
    \item \underline{Limited contextual reasoning:} Current autonomous systems cannot reason about high-level goals or modify behavior based on task complexity or environmental cues.
    \item \underline{Fault tolerance:} Unlike teleoperated systems, AUVs lack immediate human oversight, which raises concerns for risk assessment and fault recovery.
\end{itemize}
}

% \vspace{1mm}
\noindent
\rebuttal{
\textbf{Development trends}. Recent academic and industrial trends in AUVs reflect a push toward scalability, adaptability, and improved intelligence. One notable direction is multi-AUV (swarm) systems for cooperative task execution (\eg, distributed mapping, search and rescue), which significantly enhance efficiency and spatial coverage~\cite{wang2023survey}. Demand for low-cost and low-power AUVs is on the rise for academic research~\cite{jia2025research}. Additionally, learning-based techniques (\eg, LfD) are substituting explicit programming, as they allow AUVs to acquire complex skills by observing human operators~\cite{zhang2024tracking}. Finally, the fusion of DT models with AUV platforms is gaining traction for risk mitigation, allowing for virtual planning and testing of mission strategies.
}

\section{Engineering Integration of Interactive Control}\label{sec:interaction}
\begin{figure*}[t]
    \centering
    \includegraphics[width=\linewidth]{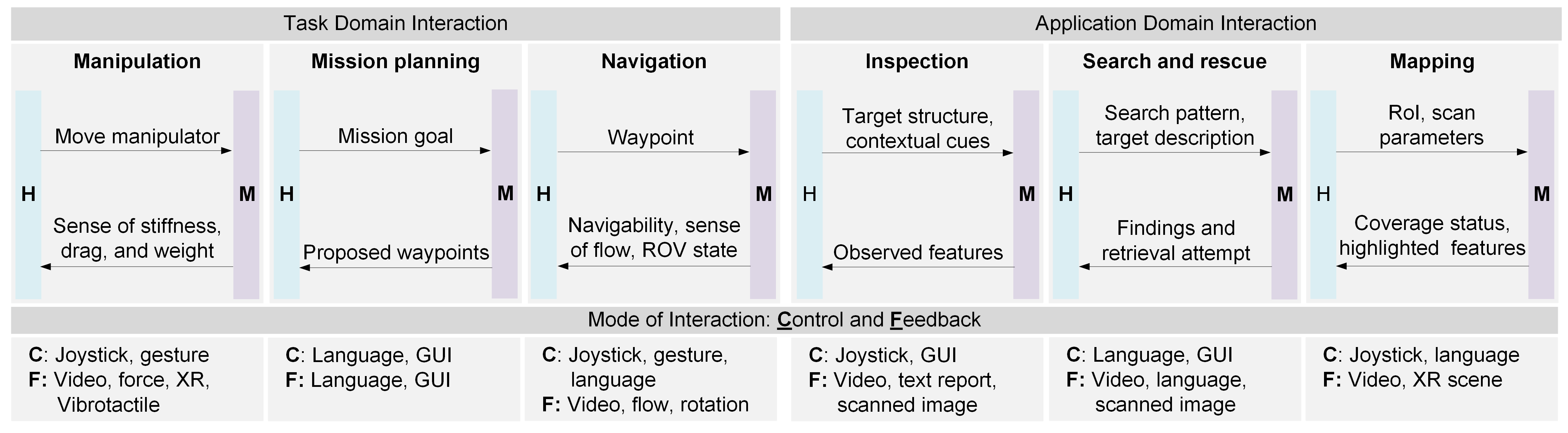}    
    \caption{Examples of bidirectional human-robot interaction in telerobotic tasks are shown, highlighting modern control and feedback mechanisms. To this end, a few prominent research topics based on \textbf{task domain} interaction are: manipulation:~\cite{ryden2013advanced,ly2021intent,kampmann2015towards,huang2022cooperative,havoutis2019learning,bruno2018augmented}, {mission planning}:~\cite{mcmahon2016mission,chen2024word2wave,yang2022digital,yang2023oceanchat,bi2023oceangpt}, {navigation}:~\cite{krishnamurthy2013self,xia2023visual,lager2018remote}; and \textbf{application domain} interaction are:  {inspection}:~\cite{boessenkool2013analysis,maurelli2016pandora,jacobi2015autonomous,grippa2022inspection,patel2024multi}, {search and rescue}:~\cite{cavallin2009semi,matos2016multiple}, and {mapping}:~\cite{fiely2024new,howard2006experiments}.}
    \label{fig:interaction}
    \vspace{-2mm}
\end{figure*}

An end-to-end telerobotic system is realized as a loop of perception, sensory feedback, and control commands. The robot first perceives the environment using multiple sensors, determines its state, and transmits the information to a surface station.
The operator interprets the remote information through various feedback mechanisms such as visual displays, auditory cues, and haptic sensations. Based on the feedback, the operator uses his cognitive intelligence to plan the motion and send control commands to the robot via switches, GUI buttons, gestures, verbal language, \etc. Maintaining this interaction loop ensures safe navigation and task completion. Fig.~\ref{fig:interaction} demonstrates some examples of human-robot interaction through command and feedback exchanges during teleoperation.

\subsection{Visual Feedback Technique}
Visual feedback is typically sent from cameras mounted inside a pressure-sealed housing, providing a front-facing first-person view of the robot. The camera sensor setup and lens type can vary widely; monocular and stereo cameras with optional IR sensors are common choices since they can provide low-light vision and depth perception. The lens can provide a wider FOV, as high as $220$\degree, representing a fisheye lens~\cite{sato2013spatio} that offers a panoramic view of the surroundings. Traditionally, the camera feed is viewed on the surface computer's screen from a fixed viewpoint; however, recent technologies such as camera arrays, HMDs, and view synthesis algorithms offer augmented peripheral views to the operator.

% \vspace{1mm}
\subsubsection{HMD}
% An HMD provides real-time video feeds from the ROV’s camera, creating a sense of being submerged in the remote environment which enhances spatial awareness and depth perception. 
% This immersive experience improves task accuracy and operator focus, especially in complex scenarios like underwater inspection, object manipulation, or navigation through confined spaces. 
In addition to displaying real-time video feed, HMDs also facilitate multi-sensor data integration, allowing operators to simultaneously monitor environmental parameters, sonar scans, and other telemetry data within the display. HMDs become even more effective when paired with a movable camera setup on the ROV~\cite{sobhani2020robot}. As the operator moves their head, the ROV's camera mimics these movements, providing a natural and intuitive way to observe the surroundings.

% \vspace{1mm}
\subsubsection{Augmented Viewpoints}
Unlike the limited first-person view from the ROV's front camera, augmented third-person perspectives offer a \textit{chase view}~\cite{hing2010development} or an overhead view, enhancing spatial awareness. 
% This expanded viewpoint enhances spatial awareness, making it easier to avoid obstacles, navigate complex environments, and plan movements more effectively. 
By toggling between first- and third-person perspectives, operators can transition from detailed, close-up control to a broader peripheral view, reducing the risk of collisions. Researchers have explored both fixed~\cite{ferland2009egocentric,lager2018remote,islam2024eob} and dynamic~\cite{nguyen2001virtual,okura2013teleoperation,abdullah2024ego2exo} viewpoint augmentation techniques, the latter being computationally expensive but providing a continuous mosaic-like view of the periphery. Augmented third-person views, when combined with real-time data overlays such as sonar or lidar maps~\cite{lager2018remote,livatino2021intuitive}, further render a rich virtual environment for interaction.

\subsection{Haptic Feedback Technique}
Haptic feedback involves sensory information derived from mechanoreceptors embedded in the skin (tactile input) and muscles, tendons, and joints (kinesthetic inputs), provided to the human operator~\cite{lederman2009haptic}. Haptic feedback in ROV teleoperation is crucial for enhancing the operator's sense of control and situational awareness, improving task performance, and reducing cognitive load during complex missions. 
The existing literature acknowledges the extremity of underwater environments and the challenges of maneuvering an ROV using a video stream alone. As such, haptic feedback is augmented into HMIs for improved control~\cite{zhou2020intuitive}. They are primarily used to warn of potential collisions and sense the ROV's interaction with the environment~\cite{ryden2013advanced}. Haptic interactions are broadly categorized into two types: kinesthetic (involves force) and tactile (involves touch and vibration).

\subsubsection{Tactile Feedback}
Tactile feedback enables humans to recognize local properties of objects being manipulated such as shapes, edges, and textures, based on measures of contact forces on the operator's skin~\cite{birznieks2001encoding}. Ideal areas for delivering this feedback include the fingertips, palm, and wrist, as they are rich in somatosensory nerves. Xia~\etal~\cite{xia2023sensory} use HaptX gloves~\cite{haptX} for a multi-level pneumatic haptic interface design\invis{; see Fig.~\ref{fig:haptic_snapshot}}. The $130$ pressure points within the user's palm and the airflow channels along the fingers make it a suitable glove to sense micro-scale turbulence as well as palm-level vibrations. Their multi-level design allows the operator to choose the appropriate level of feedback based on the task's complexity. For instance, high-resolution turbulence data is vital for accurate docking and inspection tasks, but excessive feedback during routine maintenance or exploration may overwhelm operators, leading to cognitive overload without improving efficiency~\cite{baker2020tactile}. While the feedback is engineered on land, sensing the raw information in deep water adds to the challenge since the sensors become exposed to the high pressure of the water column. Kampmann~\etal~\cite{kampmann2015towards} address this by developing SeeGrip, a rugged multi-limb manipulator capable of providing tactile sensations from deep water. To achieve a higher depth rating, a fiber-optic force sensing mechanism is proposed, unlike capacitive sensors~\cite{schmitz2011methods} or conductive rubbers~\cite{wu2010research} commonly used for terrestrial robots. 

% \begin{figure}[h]
% \centering
% \vspace{-2mm}
% \includegraphics [width=\linewidth]{images/haptic_snapshot.png}%
% \vspace{-1mm}
% \caption{A haptic teleop interface with gloves and bodysuit~\cite{xia2023sensory} is shown. 
% % \JI{@Eric: please check if fig is okay or you'd like to use something else.}
% }%
% \label{fig:haptic_snapshot}
% \vspace{-2mm}
% \end{figure}%

\subsubsection{Kinesthetic Feedback}
Kinesthetic feedback provides information on the position, velocity, force, and torque of objects through receptors in muscles and joints~\cite{edin1995skin,hayward2004haptic}, allowing operators to perceive the weight of objects or the force required to interact with them. This type of feedback is essential for precision tasks such as grasping delicate objects. Khedr~\etal~\cite{khedr2024design} design a gripper jaw that delivers force feedback to the operator's fingers, mimicking the stiffness of the grasped object. The utility of force feedback has also been explored for sensing orientation and detecting collisions. For instance, an additional hand-held steering mechanism replicating the ROV's rotation is useful, as demonstrated by Shazali~\cite{shazali2018development}. The manipulator developed by Ryden~\etal~\cite{ryden2013advanced} utilizes non-contact (camera) sensors to identify nearby obstacles and provide force feedback to the stylus controller, helping to prevent unwanted contact or collisions during manipulation. However, waterproofing bulky force sensors in high-pressure environments while maintaining sensitivity remains a significant challenge~\cite{cross2022waterproof}. Furthermore, the force-generating actuators tend to be large and power-intensive, leading to tactile feedback being the preferred choice for most applications.

% \subsection{Pseudo-Haptic Feedback}
% Given the high costs and logistical challenges of open water ROV deployment, virtual environments are beneficial for prototyping and user training. In the virtual environment, pseudo-haptic feedback creates a \textit{haptic illusion} to the operator, allowing them to perceive sensations such as friction and drag forces while observing the task from visual feedback~\cite{neupert2016pseudo}. Lecuyer~\cite{lecuyer2009simulating} utilizes pseudo-haptic feedback as a means for simulating the sensations of stiffness, texture, and mass of a virtual object through visuo-haptic perception. The virtual platforms provide operators with realistic subsea sensations~\cite{xu2021vr}, which is particularly helpful for novice operators to prepare for and minimize errors during real-world missions~\cite{xia2023rov}.

\begin{figure*}[t]
    \centering
    \includegraphics[width=\linewidth]{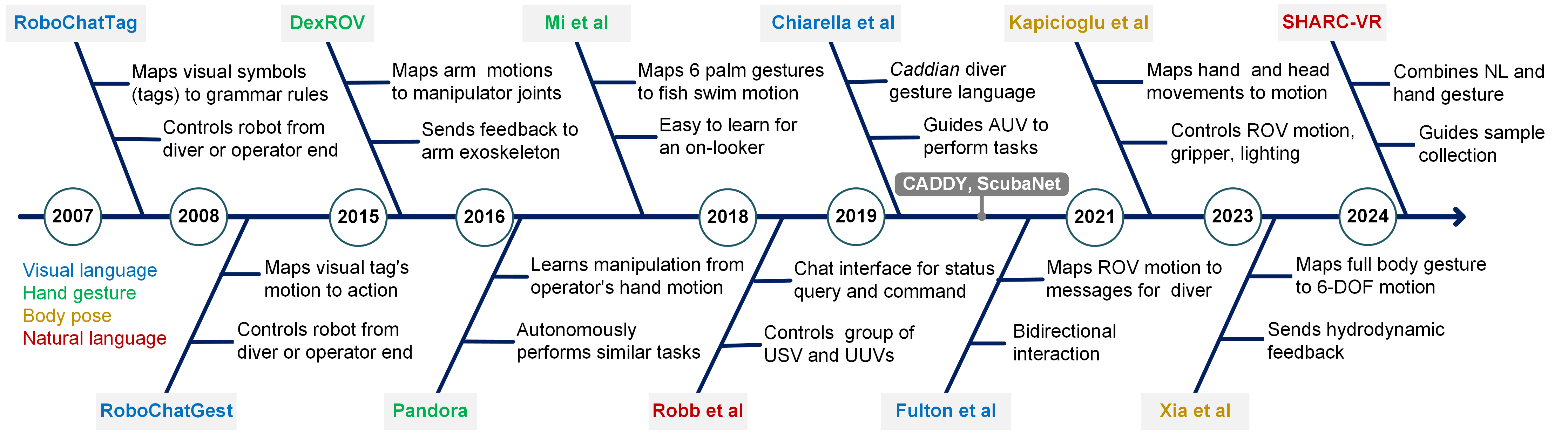}    
    \caption{A chronological progression of human-robot interaction technologies is shown, focusing on gesture-based and language-based methods. CADDY~\cite{gomez2019caddy} and ScubaNet~\cite{codd2019finding} are two comprehensive dataset for diver gesture recognition. Other key works referenced are: RoboChatTag~\cite{dudek2007visual}, RoboChatGest~\cite{xu2008natural}, DexROV~\cite{gancet2015dexrov}, Pandora~\cite{maurelli2016pandora}, Mi~\etal~\cite{mi2016gesture}, Robb~\etal~\cite{robb2018natural}, Chiarella~\etal~\cite{chiarella2018novel}, Fulton~\etal~\cite{fulton2019robot}, Kapicioglu~\etal~\cite{kapicioglu2021touchless}, Xia~\etal~\cite{xia2023sensory}, and SHARC-VR~\cite{phung2024shared}. }
    \label{fig:interaction_timeline}
    \vspace{-2mm}
\end{figure*}

\subsection{Other Feedback Cues}
In addition to visual and haptic feedback, teleoperators receive information about the robot's surroundings through other sensors. For instance, rotating sonars and lasers detect obstacles within a certain range and present the data as occupancy maps. Environmental parameters, including depth, temperature, and physicochemical properties (\eg, pH, salinity, oxygen levels), are recorded by specialized sensors and stored as mission logs or time-series data. All this information is typically displayed on a screen, overloading the pilot’s visual channel. Therefore, researchers suggest auditory cues as a means to reduce visual workload during teleoperation. The integration of auditory cues was initially explored through Audio Augmented Reality (ADAR), which combines augmented reality with a virtual audio-visual interface to facilitate improved ROV navigation ~\cite{vasilijevic2012acoustically}. However, the auditory cues used in this work are simply verbal commands (\eg, \textit{left}, \textit{right}), making it similar to NL-based interfaces. Subsequently, this approach is expanded with the development of an auditory guidance system~\cite{vasilijevic2014auditory} that utilizes non-verbal audio (\ie, auditory icons) via sonification technique~\cite{kramer2010sonification}. These works demonstrate that the concept \textit{guidance-by-sound} is feasible and more useful in noisy low-light oceanic environments~\cite{vasilijevic2018teleoperated}. Nevertheless, the authors point out that auditory feedback has much lower spatial resolution compared to vision, as humans are far more adept at visual navigation~\cite{shinn1998adapting}.

\subsection{Interactive Control}
The existing works explore various mechanisms to issue control commands to the robot, each with unique strengths depending on the task and the operational environment. Traditional joysticks and graphical user interfaces (GUI) offer a familiar and reliable method, where operators use physical controllers or software dashboards to execute specific actions. Gesture-based control systems, utilizing motion sensors and IMUs, provide an intuitive way to interact with the robot, translating the operator's hand/body movements into robotic actions~\cite{xia2023rov,maurelli2016pandora}. Lastly, NL (text/voice)-based controls introduce hands-free natural communication, allowing operators to streamline complex sequential tasks with spoken instructions~\cite{chen2024word2wave,phung2024shared}. Fig.~\ref{fig:interaction_timeline} summarizes some prominent research works on enhancing human-robot interaction via gesture-based and NL-based control modes.
% Each of these methods plays a vital role in advancing human-robot interaction, particularly in challenging underwater environments.

\subsubsection{Handheld Controller and GUI}
The traditional control method, consisting of a software console and a hardware device (\eg, joystick), remains the primary approach for teleoperating subsea robots to this day. In $1978$, NASA introduced a comprehensive command and control (C2) interface that combines a joystick, keyboard, specialized button box, and a GUI for undersea manipulation~\cite{sheridan1978human}. While these devices offer more direct, low-level control with minimal risk of interpretation errors by the robot, they demand constant attention and keep the operator's hands engaged. As a result, such systems are only suited for hands-on teleoperation with limited robotic autonomy.

\subsubsection{Hand Gesture \& Body Motion-based Control}
Utilizing hand gestures and body motion to control and communicate with underwater robots, a method extensively studied in contemporary literature, provides quick, intuitive, and instinctive responses in dynamic environments. This approach enables operators to issue commands rapidly, aligning with natural human reflexes for real-time obstacle avoidance or emergency actions. It enhances situational awareness, reduces cognitive load, and supports versatile communication, facilitating safe and efficient operations without relying on complex control interfaces~\cite{kapicioglu2021touchless,valluri2024integrating,dudek2007visual}. For instance, robots have been equipped to visually recognize divers' gestures, allowing them to follow and interact with divers~\cite{islam2018understanding,chiarella2018novel,aldhaheri2024underwater} while transmitting the feed for diver-teleoperator communication as well. Communication from robot to diver has also been tested, with robot poses/motions conveying specific information to the diver (\eg, slow looping motion indicates low battery)~\cite{fulton2019robot}. Unlike divers' gestures, understanding remote operators' gestures is more challenging since the robot receives no visual feed from the operator. Instead, the operators' gestures are captured with cameras~\cite{kapicioglu2021touchless}, body suits~\cite{xia2023rov}, wrist-mounted IMUs~\cite{babiarz2024underwater}, and actuators~\cite{xia2023sensory}, which convert the movements into electrical signals for the robot to interpret and execute. According to Camponogara~\etal~\cite{camponogara2019grasping}, haptic cues on fingers are more informative than visual feed to correct the grasping motion of a manipulator. Xia~\etal~\cite{xia2023rov} introduce a comprehensive full-body motion mapping system using the HTC OpenXR model. They assign multiple gestures to control various functions: four hand gestures manage two on/off switches, a scroll bar, and vertical movement; head rotation, detected by an HMD, adjusts roll, pitch, and yaw; body leaning, tracked by a HaptX body suit, governs horizontal motion. Their user study on $30$ novice operators shows that $90\%$ of them prefer gesture-based control over traditional joystick methods.

\subsubsection{Language Commands}

\begin{figure}[h]
    \centering
    \includegraphics[width=\linewidth]{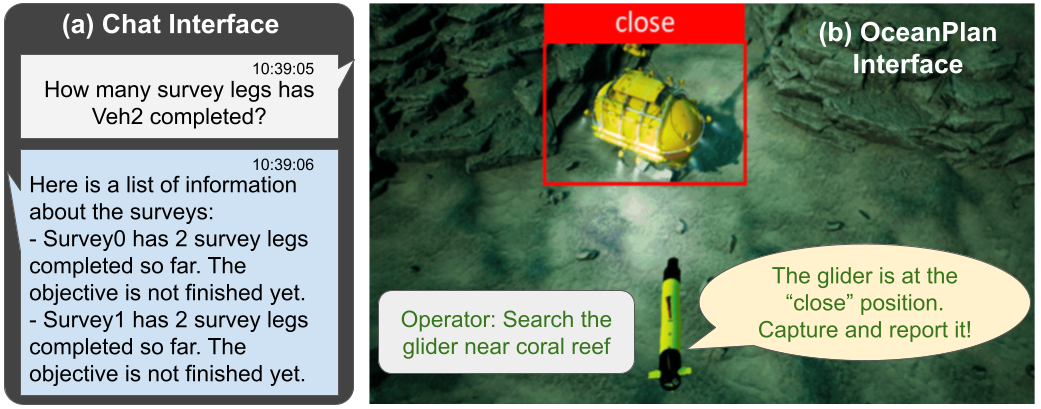}  
    \vspace{-4mm}
    \caption{Snapshots of NL-based interactions are shown; \invis{(a) waypoint mission planning from verbal command~\cite{chen2024word2wave},} (a) status query and real-time written response generation~\cite{robb2018natural}, and (b) command to execute a specific task~\cite{yang2024oceanplan}.}
    \label{fig:language_snapshot}
    \vspace{-2mm}
\end{figure}

Verbal and written NL commands represent the latest and most intuitive mode of controlling underwater robots. SOTA research showcases the potential for real-time dialogue between humans and robots, moving beyond pre-programmed commands to more flexible mission management and conversational interfaces; see Fig.~\ref{fig:language_snapshot}. Earlier interfaces such as REGIME generate post-mission summary/report from mission logs and sensory data~\cite{hastie2017talking}. MIRIAM~\cite{hastie2017miriam}, the successor of REGIME, integrates real-time mission status query options and provides important notifications such as objective completion and fault occurrences. In the Neptune autonomous command and control interface, Robb~\etal~\cite{robb2018natural} further demonstrate MIRIAM’s value in multi-robot coordination, where the operator communicates directly with the leader robot that relays information to/from the followers. For instance, when the operator asks \textit{``how many survey legs vehicle-2 has completed"}; the leader retrieves and relays that information from vehicle-2 to the operator. 
% Such control offers the flexibility of 
These interfaces operate on written instructions; hence the voice-to/from-text conversion is not thoroughly tested. More recent works utilize LLMs to address this, as demonstrated in RoboChat~\cite{li2023robochat}, OceanChat~\cite{yang2023oceanchat}, and OceanPlan~\cite{yang2024oceanplan}. These works highlight the use of verbal commands for both planning and real-time queries or adjustments. 

A comprehensive study on multi-modal hyper-redundant teleoperation highlights the task-specific advantages of both XR and language-based interfaces~\cite{martin2020application}. $94\%$ of users prefer gesture and immersive XR interfaces for grasping, object picking, and similar manipulation tasks, while they report language commands as more intuitive for navigation. Additionally, $52\%$ of the subjects favor a hybrid approach that combines verbal commands with hand gestures. Therefore, we conclude that integrating hyper-redundant control improves operator efficiency and comfort, offering more versatile solutions for diverse teleoperation tasks.

\begin{figure*}[t]
    \centering
    \includegraphics[width=0.8\linewidth]{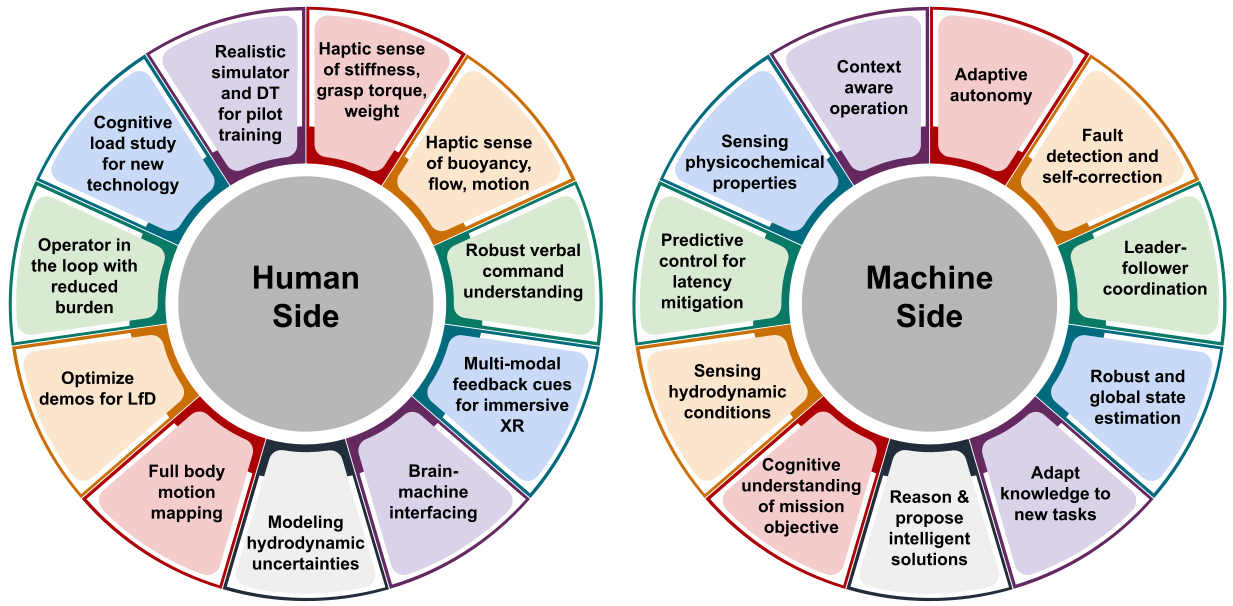}    
    \vspace{-2mm}
    \caption{Key open research problems are summarized; human-centric research focuses on improving HMIs for subsea telerobotics through immersive feedback and control, while machine-focused research prioritizes advanced sensing capabilities and intelligent decision-making.
    }
    \label{fig:open_problems}
    \vspace{-4mm}
\end{figure*}

\section{Challenges and Open Problems} \label{sec:discussion}
The subsea telerobotics domain presents unique challenges due to the hostile and dynamic nature of the underwater environment. Key technical hurdles include limited FOV, latency in long-distance operations, hydrodynamic variations, uncertainty in state estimation, and the presence of dynamic obstacles within confined spaces. Open problems in this sector span the fields of robotics, low-light perception, AI, computer vision, human cognition, and ocean engineering; see Fig.~\ref{fig:open_problems}. 
% highlights the promising directions for future research. A few prominent research areas and their open problems are discussed below.
% Addressing these challenges requires a multi-disciplinary approach, blending expertise from ocean science, robotics, AI, and human cognition to push the boundaries of subsea exploration.

% \JI{each of the following subsections need to be revised significantly, based on the guidelines I mentioned. It is understandable to feel fatigue for long papers, but these discussions are not presentable and more importantly, not very informative enough yet.}

\subsection{Human-Machine Dialogue in Natural Language}\label{subsec:NL_challenges}
As natural language has been proven to be one of the most effortless control methods for teleoperation, the future goal is to enable real-time natural conversations. LLMs, particularly GPT models~\cite{he2023robotgpt}, have successfully conducted real-time human-machine dialogues for various HRI settings~\cite{zhang2023large,huang2022inner}. However, marine robots face the additional challenges of integrating sensory data into their decision-making process while executing an operator's command. ChatSim~\cite{palnitkar2023chatsim} and OceanGPT~\cite{bi2023oceangpt} have shown some progress in this regard, demonstrating limited dialogue capabilities and gathering vast knowledge of the ocean environment into the model, respectively. Further research can enable robots to reason logically and propose alternative mission strategies.

% \JI{the second para needs to be connected to the conversation. Seems something else all of a sudden}
While language models can parse instructions and reason from a long conversation, they struggle with linguistic biases~\cite{fleisig2024linguistic} such as variations in dialect/pronunciation, indirect commands, as well as mission-specific terminologies. Therefore, another research direction would be a risk assessment for NL interfaces and developing error-handling mechanisms for the robot to identify, query, and correct ambiguous or potentially dangerous commands. \rebuttal{In the natural language generation (NLG) domain, errors are corrected via human feedback and reinforcement learning~\cite{izadi2024error}. Lin~\etal~\cite{lin2021using} compare three error handling strategies such as reprompt, suggestion, and confirmation, across users of different age groups interacting with a chatbot. Among telerobotic NL interfaces, ChatSim~\cite{palnitkar2023chatsim} utilize closed perception-action loops to minimize errors in GPT-generated mission plans. However, other interfaces are yet to adopt any structured error-handling mechanisms. To clarify ambiguous user queries, researchers use Similar Question Model (SQM)~\cite{trienes2019identifying} that identifies ambiguous utterances by extracting features from past similar queries. Other approaches to reduce dialect and linguistic errors include phonetic normalization, code mixing (\ie, mixing words from multiple languages)~\cite{joshi2025natural} \etc. To this end, these correction methods have been applied to search agents and AI assistants~\cite{keyvan2022approach}; hence, their utility for robotic interfaces remains underexplored.}
% to this end, ambiguous command correction has not been extensively investigated  and leaves scope for future development. 

% the interface asks follow-up questions

% \reviewer{(\#13)
% \begin{itemize}
%     \item Could you expand on the error handling mechanisms employed in existing NL interfaces for subsea telerobotics? It sounds like this might be a critical problem.
%     \item Following up, how do these systems address issues such as variations in dialect, tone, or pronunciation?
%     \item Again, this section can easily be expanded with additional explanations. Are there any strategies for providing feedback to the operator when a command is ambiguous or potentially dangerous? Some methods could be indicated here.
% \end{itemize}
% }

% \begin{itemize}[left=0pt]

%     \item More research on language models is required to achieve the next level of intelligence, where the robot would maintain and process the mission context throughout a long, complex dialogue exchange with the operator.

%     \item Another aspect of intelligence would be developing cognitive understanding; the robot will be able to not only interpret commands but also reason and propose alternative strategies during the mission.

%     \item Novel fail-safe mechanisms are necessary for the robot to identify, query, and correct potentially dangerous commands that may lead to system failure.

    % \item The robot’s ability to interpret nuanced languages, such as indirect commands, questions, and mission-specific terminology requires further improvement.

% \end{itemize}

\subsection{Optimizing Learning from Demonstration (LfD)}
The robot's ability to learn from repetitive tasks and apply that knowledge to unknown scenarios remains under-explored in marine robotics, unlike terrestrial and aerial domains~\cite{ravichandar2020recent}. Few recent works have tested LfD for underwater manipulation~\cite{havoutis2019learning,maurelli2016pandora}, and sample collection~\cite{somers2016human}. However, extensive real-world evaluation is necessary to adopt these developments into industrial applications.

In general, reinforcement learning (RL) pipelines struggle with dynamic changes such as a robot's component failure or unseen terrain~\cite{nagabandi2018learning}. Extensive training (\ie, demonstrations) in diverse scenarios is a straightforward solution. However, crowdsourcing demonstrations is extremely challenging in harsh subsea environments, unlike ground/aerial scenarios~\cite{ravichandar2020recent}. This motivates multiple research directions. First, the robot should \textit{learn how to adapt} to an unseen task using prior experience. Meta RL~\cite{finn2017model} and online RL~\cite{nagabandi2018learning} have shown progress in such model adaptation through reward-driven optimization. Researchers have utilized these approaches for improving subsea locomotion~\cite{li2021deep} and path planning~\cite{yang2022intelligent}. Second, realistic simulators featuring various scenes can be a potential alternative for recording numerous demonstrations. Third, algorithmic studies can reveal the minimum number of demonstrations required to achieve a desired performance threshold. Additionally, hybrid approaches (LfD and RL) should be explored, where operator interventions during autonomous missions are treated as demonstrations or corrective feedback. Such interventions could potentially \textit{teach} the robot to generalize knowledge and refine policies. %, and avoid similar errors in the future.

% exploring how an operator's intervention can be insightful for a robot during the learning process is essential. Such insights could enable the robot to generalize and transfer knowledge, facilitating adjustments to its autonomous actions in future tasks.

% LfD is useful for maintenance and inspection tasks in unpredictable or hazardous subsea conditions. Further research on LfD could accelerate navigation performance as well, reducing the need for extensive retraining when the robot faces unknown structures/scenes. Several challenges and open problems of this technology are yet to be addressed.

% \begin{itemize}[left=0pt]
%     \item 

%     \item Training efficiency would be another key factor; extensive study on learning algorithms can optimize the number of demonstrations required for a robot to master complex tasks.
%     % \item Developing algorithms that can generalize the knowledge base and translate it toward new operations is a major scientific problem to solve. 

%     \item It is essential to explore how operator feedback can be integrated into the learning process to refine and adjust robotic actions in real-time during underwater operations.
    
% \end{itemize}

% However,  and developing algorithms that can effectively translate a robust knowledge base for other operations. 

\subsection{Hydrodynamic Sensing \&  Haptic Feedback Modeling}\label{subsec:hydro_haptic}

\finalrev{A real-time hydrodynamic-to-haptic pipeline requires: \textbf{(i)} onboard flow or force sensing, \textbf{(ii)} estimation of operator-relevant hydrodynamic loads, and \textbf{(iii)} stable haptic rendering. Doppler current profilers and artificial lateral-line sensors can measure far- and near-field flow, respectively~\cite{xia2023rov}. High-fidelity SPH and PBD are primarily suitable for offline simulation and training-data generation~\cite{liu2010smoothed,macklin2013position}; real-time feedback instead requires reduced-order or physics-informed surrogates driven by live measurements~\cite{cai2021physics}. Generative models may augment rare training scenarios but should not directly drive haptic cues without uncertainty validation~\cite{drygala2022generative}. In this domain, open problems include bounding latency and model error, mapping complex flows into perceptually useful and ergonomically safe cues, and validating their effects on workload and task performance.}

\subsection{Latency Mitigation for Long Distance Operation}

\finalrev{\textbf{Communication-level mitigation} seeks to improve the availability and capacity of the link through LAN or satellite relays, optical-acoustic channels, RF communication, and compact magnetoelectric antennas~\cite{lachaud2018opportunities,gancet2015dexrov,farr2010integrated,sani2018fitoplankton,talebi2024blueme}. Their practical limitations include range, throughput, alignment, environmental sensitivity, energy consumption, and reliability. Open problems include seamless handoff between communication modes and resource-aware transmission that prioritizes safety-critical telemetry when bandwidth or energy deteriorates~\cite{sendra2015underwater,song2013efficient}.}

\finalrev{\textbf{Perception-level mitigation} uses frame interpolation and predictive displays to preserve situational awareness during short feedback gaps~\cite{lotter2016deep,rank2016predictive}. However, predicted imagery may conceal unexpected obstacles or present an incorrect future state. Such interfaces must distinguish measured from predicted information, communicate uncertainty, and determine when a prediction is no longer trustworthy. Their effects on operator trust, collision risk, and task accuracy also require validation under realistic delay conditions.}

\finalrev{\textbf{Autonomy-level mitigation} addresses the effects of delay on command execution. Passivity-based or model-mediated control can preserve stability or responsiveness during shorter delays, while supervisory, shared, or local autonomy can maintain stabilization and safety functions during longer communication loss~\cite{anderson1991stable,mitra2008model,gancet2015dexrov,du2024sensory}. Remaining challenges include command arbitration, safe authority transitions, operator mode awareness, and reconciliation between delayed operator commands and autonomous actions after reconnection. No individual layer eliminates delay; therefore, robust systems must coordinate all three levels and degrade safely as communication quality changes.}

\subsection{Real-World to/from Simulation}
Existing virtual simulators are capable of rendering simplified models of underwater workspace, including surface waves~\cite{paleblue}, chemical profiles~\cite{zhang2024uuvsim}, coral reefs~\cite{m8rsim}, subsea rigs~\cite{abyssal}, and some dynamic obstacles~\cite{amer2023unav}. However, they fail to represent the turbulent oceanic conditions and waterbody characteristics~\cite{siddique2024aquafuse} in general. As compared earlier in Table~\ref{tab:simulator}, only a few simulators (\eg, UWSim and UUV Simulator) provide an extensive choice of sensors and diverse scenarios to support navigation, SLAM, and planning algorithms. Many platforms cannot simulate interactions with dynamic marine objects, which is critical for the aforementioned tasks. The utility comparison of various simulators~\cite{ciuccoli2024underwater} also highlights these limitations. Expanding the sensor suites and scenario diversity in simulators remains an open research area.

Another limitation of existing simulators is their inability to incorporate real-world data, leaving a gap between simulation and real-world missions. Robust \textit{real2sim} frameworks are needed to translate real-time and historical sensor data into virtual environments. For instance, project RUMI~\cite{fiely2024new} leverages real-time video feeds to create an interactive terrain within a digital environment to improve realism~\cite{lim2022real2sim2real}. Eventually, this approach may facilitate \textit{sim2real} transfer of algorithms developed in simulation~\cite{zhang2022sim2real}. Moreover, many digital shadow/twin models fail to accurately replicate real-world pairs, lacking fidelity in sensor modeling and environmental dynamics~\cite{ciuccoli2024underwater}. 

\finalrev{These fidelity gaps limit how confidently simulator results can be transferred to field operation. Simulation-based HMI evaluation should therefore be complemented by physical trials that expose the system and operator to uncertain visibility, currents, communication degradation, hardware faults, and operational stress. For training applications, evaluation should measure both immediate performance and delayed skill retention across repeated and long-duration sessions. Scenario diversity and progressive difficulty reduce overfitting to a particular simulator, while periodic physical trials verify that learned procedures transfer to real missions. Relevant measures include task accuracy, completion time, error recovery, workload, and fatigue.}

\section{Conclusions} \label{sec:conclusion}
The increasing dependence on ROVs and AUVs for subsea operations underscores the critical need for advanced HMIs that enable precise control, robust interaction, and mission success. \finalrev{In this review, we organized subsea HMI research around three interrelated perspectives: operator experience, robotic autonomy, and bidirectional human-machine communication.} We identified the unique operational challenges inherent to subsea teleoperation and examined state-of-the-art approaches designed to mitigate them. \finalrev{We compared interface modalities, simulators, digital twins, and shared-autonomy approaches while distinguishing established industrial functions from selectively deployed and research-stage capabilities. The synthesis reveals persistent gaps in interoperability, communication reliability, immersive feedback, and field validation, and clarifies that simulation-based evidence must be complemented by physical evaluation. These findings motivate research on hydrodynamic feedback, latency-aware shared autonomy, grounded natural-language interaction, standardized evaluation, and reliable sim-to-real or real-to-sim transfer.} Addressing these challenges will require sustained interdisciplinary collaboration across robotics, artificial intelligence, interfacing systems, and cognitive science to advance the next generation of HMI systems for subsea telerobotics.

\section*{Acknowledgments}
This work is supported in part by the National Science Foundation (NSF) grants \#$2330416$ and \#$2326159$, and the University of Florida (UF) research grant \#$132763$.

% In this survey, we explore various technical aspects of human-machine interface technologies for subsea robotic operations. We summarize the historical evolution and scientific breakthroughs in this domain and assess their impact on reducing operator's cognitive load and enhancing mission safety. Particularly, we provide extensive literature review on traditional vision-based teleoperation consoles, and more advanced interfaces with haptics, XR, and NL integration. We discuss the virtual interfaces and digital ROVs as a cost-effective alternative for training and initial testing purposes. We identify the challenges of achieving autonomy in harsh dynamic underwater conditions, and investigate various shared autonomy mechanisms for interfacing with intelligent robots. Moreover, we highlight the SOTA feedback and control mechanisms for immersive interaction with the robot and the remote environment. Finally, we highlight the exciting open problems in this domain and predict the future technologies of this domain. 

% ---------- References ----------
{\footnotesize
\bibliographystyle{plain}
\bibliography{bibs/ICIC_refs,bibs/robopi_pubs,bibs/reference}
}

\end{document}